\documentclass[letterpaper,journal]{IEEEtran}
\usepackage{amsmath,amsfonts}
\usepackage{algorithm}
\usepackage{algorithmic}
\usepackage{array}
\usepackage{textcomp}
\usepackage{stfloats}
\usepackage{url}
\usepackage{verbatim}
\usepackage{graphicx} 
\usepackage{tabularx}
\usepackage{booktabs}
\usepackage{stfloats}
\usepackage{multirow} 
\usepackage[T1]{fontenc}
\usepackage{booktabs}
\usepackage{bigstrut}
\usepackage{amsmath}
\usepackage{graphicx}
\usepackage{epstopdf}
\usepackage{CJKutf8}
\usepackage{booktabs}    
\usepackage[section]{placeins}
\usepackage{float}
\usepackage[numbers,sort&compress]{natbib}
\usepackage{threeparttable}  
\usepackage[colorlinks,bookmarksopen,bookmarksnumbered,citecolor=blue, linkcolor=blue, urlcolor=blue]{hyperref}

\usepackage {upgreek}
\usepackage {bm}

\RequirePackage[format=plain,labelformat=simple,labelsep=period,font=small,compatibility=false]{caption}
\RequirePackage[font=footnotesize,skip=3pt,subrefformat=parens]{subcaption}

\usepackage[numbers,sort&compress]{natbib}
\def\BibTeX{{\rm B\kern-.05em{\sc i\kern-.025em b}\kern-.08em
		T\kern-.1667em\lower.7ex\hbox{E}\kern-.125emX}}

\begin{document}
	
	\title{Event-based high temporal resolution measurement of shock wave motion field}
	
	\author{Taihang Lei, Banglei Guan, Minzu Liang, Pengju Sun, Jing Tao, Yang Shang, and Qifeng Yu \vspace{-0.56cm}
	
	\thanks{Manuscript received ** *, 2025; revised ** *, 2025; accepted * *, 2025. Date of publication * *, 2025; date of current version * *, *. This work was supported by the National Natural Science Foundation of China (Grant No. 12372189) and the Hunan Provincial Natural Science Foundation for Excellent Young Scholars (Grant No. 2023JJ20045) and the Science and Technology Innovation Program of Hunan Province (Grant No. 2022RC1196). The Associate Editor coordinating the review process was ** *. (Corresponding authors: Banglei Guan.)(e-mail: guanbanglei12@nudt.edu.cn).}
	\thanks{Taihang Lei, Banglei Guan, Pengju Sun, Jing Tao, Yang Shang, and Qifeng Yu are with the College of Aerospace Science and Engineering, National University of Defense Technology, Changsha 410073, Hunan, China, and with the Hunan Provincial Key Laboratory of Image Measurement and Vision Navigation, Changsha 410073, Hunan, China.
    
    Minzu Liang is with the College of Science, National University of Defense Technology, Changsha 410073, Hunan, China.}}

\markboth{IEEE Transactions on Circuits and Systems for Video Technology,~Vol.~XX, No.~X, XXXX~2025 }%
{Shell \MakeLowercase{\textit{et al.}}: A Sample Article Using IEEEtran.cls for IEEE Journals}

\maketitle

\begin{abstract}
Accurate measurement of shock wave motion parameters with high spatiotemporal resolution is essential for applications such as power field testing and damage assessment. However, significant challenges are posed by the fast, uneven propagation of shock waves and unstable testing conditions. To address these challenges, a novel framework is proposed that utilizes multiple event cameras to estimate the asymmetry of shock waves, leveraging its high-speed and high-dynamic range capabilities. Initially, a polar coordinate system is established, which encodes events to reveal shock wave propagation patterns, with adaptive region-of-interest (ROI) extraction through event offset calculations. Subsequently, shock wave front events are extracted using iterative slope analysis, exploiting the continuity of velocity changes. Finally, the geometric model of events and shock wave motion parameters is derived according to event-based optical imaging model, along with the 3D reconstruction model. Through the above process, multi-angle shock wave measurement, motion field reconstruction, and explosive equivalence inversion are achieved. The results of the speed measurement are compared with those of the pressure sensors and the empirical formula, revealing a maximum error of $\textbf{5.20\%}$ and a minimum error of $\textbf{0.06\%}$. The experimental results demonstrate that our method achieves high-precision measurement of the shock wave motion field with both high spatial and temporal resolution, representing significant progress. 
\end{abstract}


\footnote{Copyright © 20xx IEEE. Personal use of this material is permitted. However, permission to use this material for any other purposes must be obtained from the IEEE by sending an email to pubs-permissions@ieee.org.}

\section{Introduction}
\label{sec:introduction}
\IEEEPARstart{S}{hock} waves are significant phenomena generated by explosions, fundamentally characterized as disturbances\cite{ref1}. When a warhead detonates, the high-temperature and high-pressure products expand rapidly, creating shock waves in the surrounding medium. Shock waves generated by the warhead are a major cause of damage\cite{ref2}. Understanding their propagation mechanisms and destructive effects is essential for equipment development and engineering protection\cite{ref3}. In particular, accurately measuring shock wave motion parameters with high spatiotemporal resolution provides critical data for power field testing and damage assessment. However, the fast and uneven propagation speed of shock waves and the unstable testing environment present significant challenges in measurement. Therefore, there is an urgent need to develop methods for measuring shock wave motion fields with high spatiotemporal resolution, strong flexibility, and the capability to estimate propagation asymmetry.


Currently, common measurement methods for explosion shock waves include electrical measurement, the effect target method, and high-speed photography\cite{ref4}. The electrical measurement method captures overpressure data by placing pressure sensors in the explosive power field and can be classified into lead testing and storage testing approaches\cite{ref5}. Due to its high accuracy and stability, the electrical measurement method remains the predominant technique for shock wave assessments. In contrast, the effect target method estimates the overpressure characteristics by analyzing the deformation of targets post-experiment. However, these methods pose challenges for the reliable quantitative analysis of shock waves\cite{ref6}. The measurement methods mentioned earlier use pre-set points. This results in discrete observation data, which limits their effectiveness in modeling shock wave propagation. As a novel power field testing technique, high-speed photography garners considerable attention in recent years. It offers significant advantages, including strong flexibility\cite{ref35}, continuous observation\cite{ref36}, and the ability to conduct repeatable experiments\cite{ref37}\cite{ref7}. However, high-speed cameras face the challenge of balancing high dynamic range with high temporal resolution, particularly under the harsh lighting conditions typical of explosion scenarios.

In recent years, new types of visual sensors have emerged in neuromorphic vision\cite{ref8}\cite{ref39}. Those cameras retain the traditional advantages of optical measurement while also offering benefits such as high dynamic range\cite{ref38}, low latency\cite{ref9}, and low cost\cite{ref10}. One such sensor is the event camera, which, due to its high dynamic range, can directly capture images containing the blast center when observing shock waves. Its low latency also enables the recording of shock waves with microsecond-level time resolution. Additionally, by recording only changes, event cameras significantly reduce irrelevant information during the observation process, enhancing both temporal resolution and sustainability.

In this paper, multiple event cameras are employed to observe shock waves and develop a novel framework. This framework enables the handling of microsecond-level timestamp events, high-temporal-resolution multi-angle shock wave measurement, 3D motion field reconstruction, and explosive equivalence inversion. The key innovative aspects are as follows:

\begin{itemize}
	\item[$\bullet$] A polar coordinate encoding is proposed to model the radial motion and propagation angle of shock waves, utilizing information from the blast center. An adaptive ROI extraction algorithm, based on event offset calculations, efficiently extracts candidate events for each angle.
	\item[$\bullet$] A slope-iterative algorithm for extracting shock waves from the ROI is proposed, leveraging the continuity of shock wave velocity changes. By preserving microsecond-level temporal resolution of events, this approach enables precise shock wave measurements.
	\item[$\bullet$] Based on the event-based optical imaging and geometric projection models, universal radius solution and 3D reconstruction models are developed. These enable shock wave asymmetry estimation and motion field reconstruction.
\end{itemize}

\section{Related work}
In this chapter, we first examine the hardware foundations and applications of event cameras, followed by a review of relevant research on optical measurements of shock waves.

\subsection{Event-based vision}
The concept of the event camera originated in 1992\cite{ref11}, and its hardware performance has significantly improved over the past thirty years\cite{ref44}. Major emerging manufacturers of these cameras include celePixel\cite{ref12}, prophesee\cite{ref13}, iniVation\cite{ref14}, and others. Currently, event cameras are employed as optical sensors in various visual tasks, yielding promising results. In the realm of 3D reconstruction and depth estimation, Muglikar et al.\cite{ref15} employ event cameras to address Shape-from-Polarization, successfully balancing accuracy with temporal resolution. Klenk et al.\cite{ref16} exploit the high dynamic range and no-motion-blur characteristics of event cameras to estimate neural radiance fields from rapidly moving sources. Liu et al.\cite{ref40} utilize event camera and introduced the Recurrent Transformer for monocular depth estimation, while Uddin et al.\cite{ref45} focus on stereo depth estimation. In Simultaneous Localization And Mapping (SLAM) tasks, Fischer et al.\cite{ref17} leverage the asynchronous output and high temporal resolution of event cameras, facilitating rapid location tracking. Guo et al.\cite{ref18} utilize event cameras for rotational motion, implementing SLAM through Contrast Maximization techniques. In the area of visual tracking, both Bonazzi et al.\cite{ref19} and Chen et al.\cite{ref20} take advantage of the low power consumption and high temporal resolution of event cameras to address eye tracking challenges. In recent research, the high-speed and high dynamic range advantages of event cameras are further explored. Xu et al.\cite{ref46} focus on utilizing the high-speed advantage of events to remove image blur, while Bisulco et al.\cite{ref47} focuses on applications under low light conditions. Collectively, these studies demonstrate that the unique advantages of event cameras, render them particularly suitable for high-dynamic range\cite{ref42}\cite{ref48}, high-speed\cite{ref41}\cite{ref49}, and rapidly changing scenes\cite{ref43}.

\subsection{Optical measurement of shock waves}
The optical measurement method for shock waves has developed relatively recently. However, due to its advantages in continuity and flexibility of observation, several research studies have emerged in recent years. The most commonly used optical sensor is the high-speed visible light camera, which faces the challenge of balancing high dynamic range with high temporal resolution. Furthermore, high-speed cameras capable of extremely high frame rates often cannot store data continuously for extended periods, leading to increased observation costs.

The most relevant work to our study is that of Kyle et al.\cite{ref21}, who utilizes multiple high-speed cameras to estimate the asymmetry of conventional blast shock waves. However, their work focuses on synchronous image frames, while ours addresses asynchronous event streams. Furthermore, the proportional measurement model used in their research may introduce significant errors when the shock wave radius is large. Additionally, we also estimate the velocity of the shock wave. Xu et al.\cite{ref22} employ a high-speed camera to observe both the initial and reflected waves. Nevertheless, this approach offers only a single observational view and is primarily focused on engineering applications. Some studies utilize pipelines to facilitate the observation of shock waves\cite{ref23}\cite{ref24}, thereby simplifying the experimental process. However, this approach may limit the ability to assess the asymmetry of the shock waves. Other studies focus on distinct application scenarios, such as Higham et al.\cite{ref25}, Rigby et al.\cite{ref26} and Gomez et al.\cite{ref27} research on near-field shock waves, while Li et al.\cite{ref28} and Zhang et al.\cite{ref29} investigate underwater shock waves.

In our previous work\cite{ref30}, we utilized a monocular event camera to observe shock waves generated by conventional explosions propagating along the ground. However, the requirement to accumulate event frames during processing resulted in a loss of temporal resolution. Furthermore, this approach is unable to estimate any asymmetry of shock wave propagation. In contrast, the new method presented in this article overcomes these limitations.

\section{Event-based measurements of shock waves}
Event cameras are bio-mimetic optical sensors that differ from traditional imaging systems. Instead of capturing images at fixed intervals, event cameras utilize a unique logarithmic logic circuit to continuously monitor light intensity across each pixel at very brief intervals $\xi$(as low as $1\upmu\mathrm{s}$). An event, represented as $e_s(x_s,y_s,p_s,t_s)$, is generated only when the change in light intensity within a pixel exceeds a predetermined threshold $\Phi$. The brightness change $\Delta L$ at pixel $(x_s,y_s)$ at time $t_s$ is expressed by:
	\begin{eqnarray}
              \Delta L=L(x_s,y_s,t_s)-L(x_s,y_s,t_s-\xi)
        \end{eqnarray}
then the polarity $p_s$ is defined by the following equation:
	\begin{align}
              \left\{
              \begin{aligned}
                   p_s&=+1, &if \quad \Delta L \ge 0 \quad and \quad \lvert \Delta L \rvert \geq \Phi\\
                   p_s&=-1, &if \quad \Delta L \le 0 \quad and \quad \lvert \Delta L \rvert \geq \Phi
              \end{aligned}
              \right.
        \end{align} 
Each pixel works independently, resulting in asynchronous output events. This characteristic enables the event camera to perform microsecond-level observations.

Our framework to measure shock waves using event cameras is illustrated in Fig. 1. First, the two-dimensional image coordinates of the blast center are utilized as reference points to convert the observed events $e_s(x_s,y_s,p_s,t_s)$ into polar coordinates $e_s(d_s,\alpha_s,p_s,t_s)$. From the distance $d$ and time $t$ of all events, a $d-t$ diagram is generated to reveal the characteristics of the shock wave front. Building on this foundation, an adaptive ROI extraction algorithm is proposed to handle events of each angle efficiently and automatically. Next, the distribution density of events within the ROI and the continuous propagation characteristics of shock waves are considered. A slope-iterative algorithm is then employed to extract microsecond-level events associated with the shock wave front. Subsequently, a general geometric model of the shock wave radius and events is derived from these extracted data. By integrating the intrinsic and extrinsic parameters of the event camera system, measurements of shock wave motion parameters from multiple views and propagation angles are achieved. This framework estimates the asymmetry of the shock wave to some extend and leads to the 3D reconstruction of the shock wave motion field. Furthermore, equivalent of the charge is calculated using the measured parameters.

\begin{figure*}
  \centering
  \includegraphics[width=0.98\textwidth,height=!]{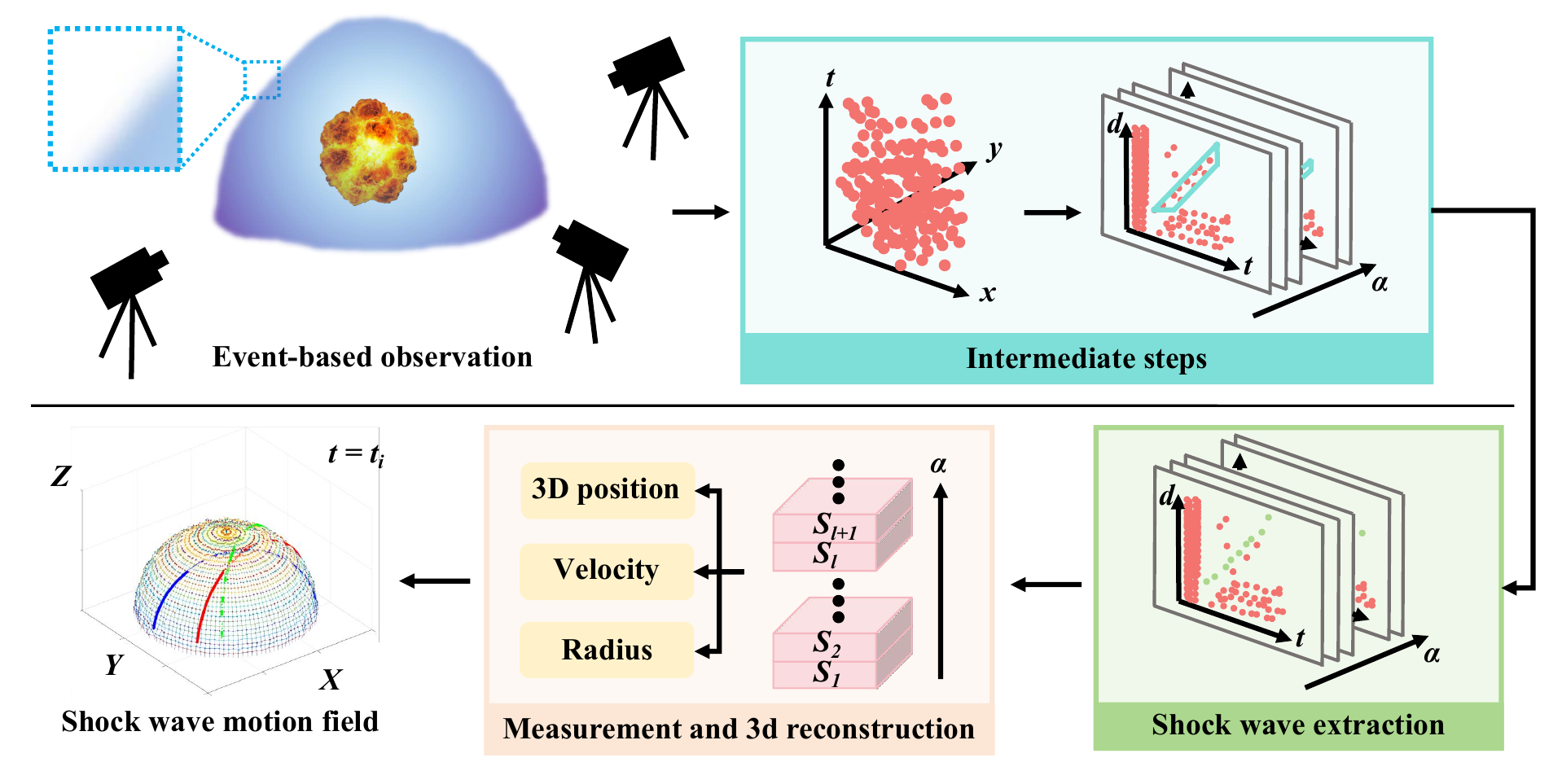}
  \caption{Overview of the proposed framework. By processing the event stamp through three modules, the asymmetry of the shock wave is estimated, and the motion field is reconstructed.}
\end{figure*}

\subsection{LED-based event cameras system calibration}

In our work, a strobe LED marker combined with RTK (Real-Time Kinematic) positioning are utilized to calibrate the event camera observation system. Additionally, a simple and efficient LED marker extraction algorithm is developed for this purpose. 
\begin{figure}[h]
  \centering
  \includegraphics[width=0.48\textwidth,height=!]{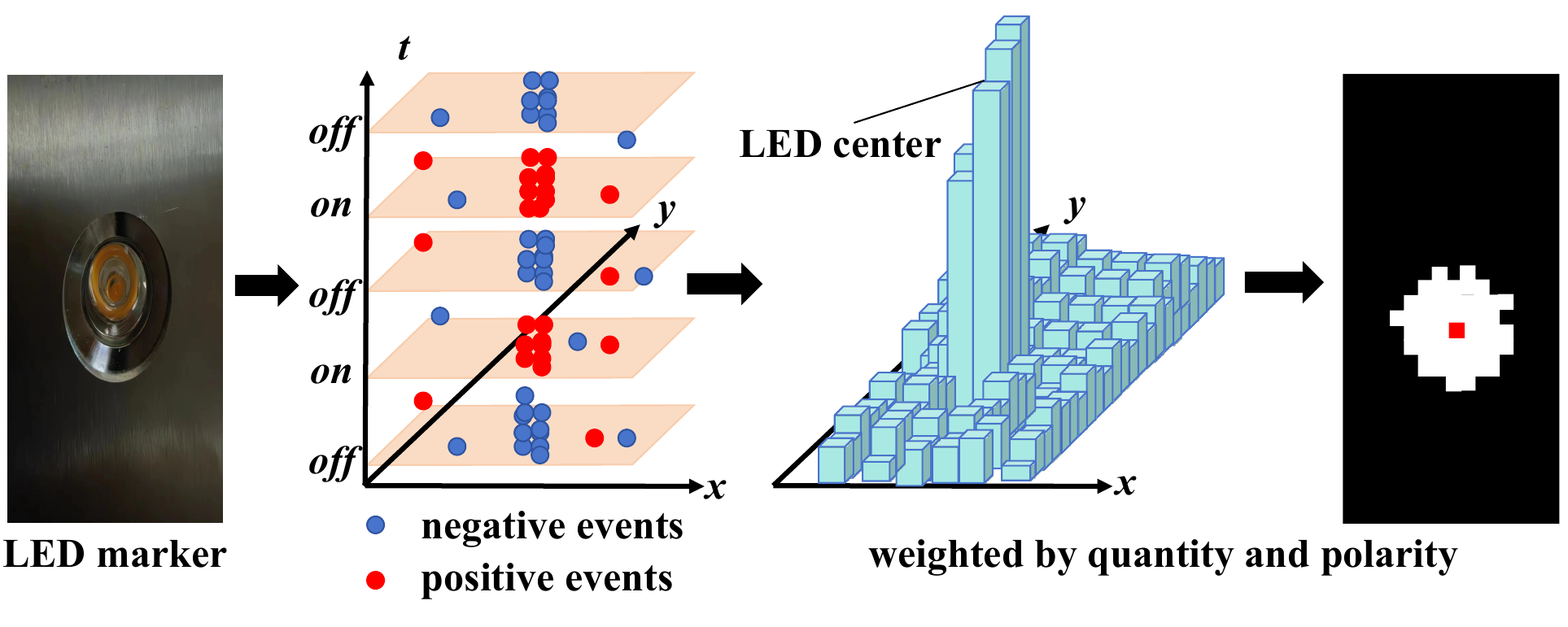}
  \caption{Extraction process of LED marker. The density and polarity information of each pixel triggering event are used for extraction.}
\end{figure}

The events generated by LED signs exhibit significant spatial density distribution and distinct polarity characteristics within the event stamps. To analyze this, we first count the total number of triggers per pixel within a very short time window (one flicker cycle) to create a statistical matrix $D$. Subsequently, the coarse positioning of the LED marker is determined based on:
	\begin{eqnarray}
              \left\{
              \begin{array}{cc}
                   x'=\mathop{\max}\limits_{i} D_{ij}\\
                   y'=\mathop{\max}\limits_{j} D_{ij}
              \end{array}
              \right.
        \end{eqnarray} 
where $i$ and $j$ are index of coordinates, $x'$ and $y'$ are rough extraction coordinates of LED marker. Afterwards, a search window is established around the coarse-extracted pixels of the LED marker. Based on the principle that the number of positive and negative polarity events influenced by the flicker should be equal, all pixels are weighted to calculate the centroid according to:
	\begin{align}
              \left\{
              \begin{array}{l}
              \displaystyle
                   x''=\frac{\sum_{i=x'-q}^{x'+q}\sum_{j=y'-q}^{y'+q}iD'_{ij}}{\sum_{i=x'-q}^{x'+q}\sum_{j=y'-q}^{y'+q}D'_{ij}}\\
                   \displaystyle
                   y''=\frac{\sum_{i=x'-q}^{x'+q}\sum_{j=y'-q}^{y'+q}jD'_{ij}}{\sum_{i=x'-q}^{x'+q}\sum_{j=y'-q}^{y'+q}D'_{ij}}\\
                   D'_{ij}={\min}(D^+_{ij},D^-_{ij})
              \end{array}
              \right.
        \end{align} 
where $D^+_{ij}$ represents the statistical matrix of positive polarity events, $D^-_{ij}$ denotes the statistical matrix of negative polarity events, $q$ is the size of search window, and $(x'',y'')$ refers to the precise extraction results of the LED marker. The entire extraction process is shown in the Fig. 2.

\subsection{Intermediate steps in shock wave extraction}

In this subsection, intermediate steps including polar coordinate encoding, d-t diagram formation, and adaptive ROI extraction are proposed, as illustrated in Fig. 3. The process takes an event stream as input and outputs the candidate events triggered by the shock wave front. 
\begin{figure}[h]
  \centering
  \includegraphics[width=0.48\textwidth,height=!]{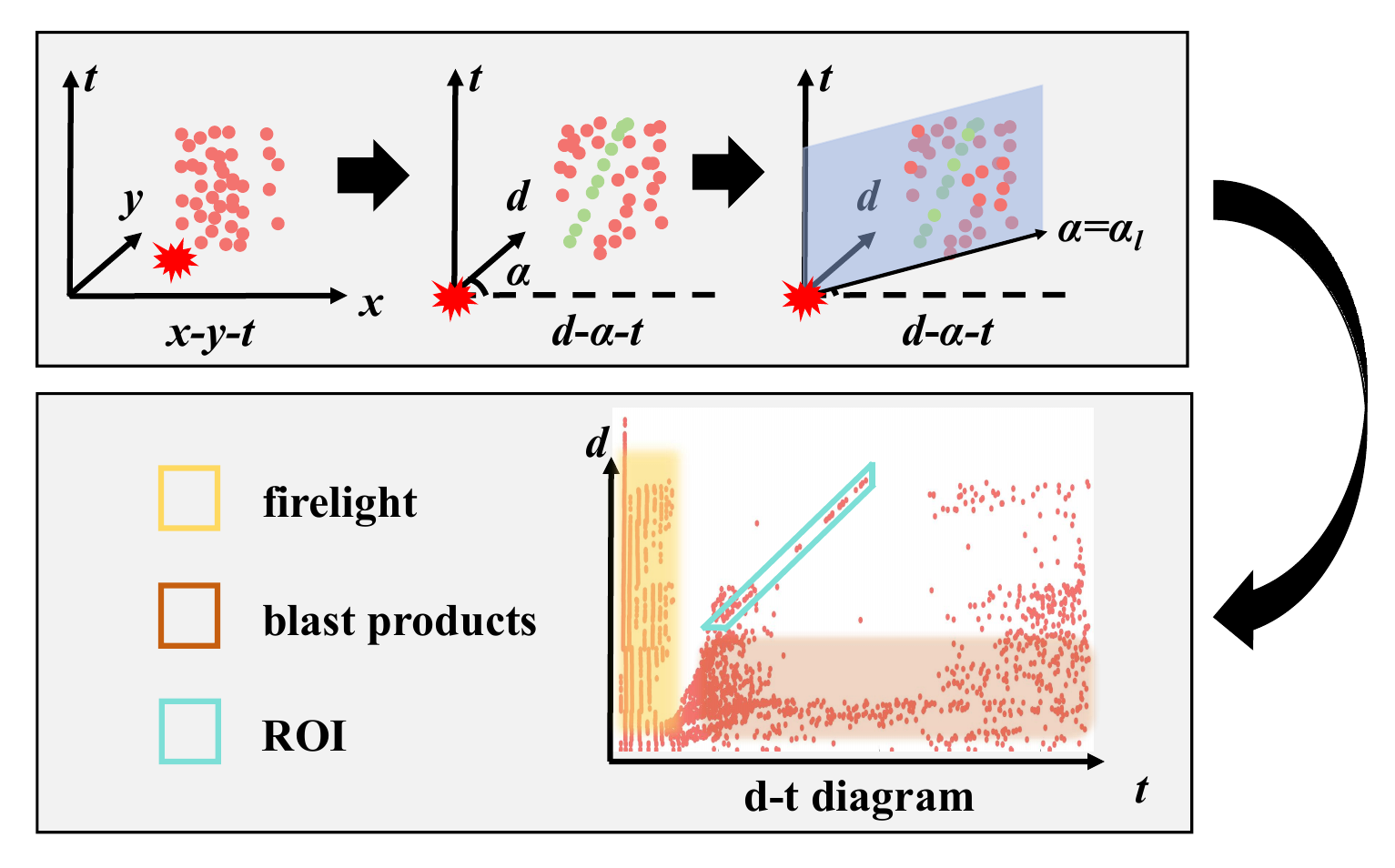}
  \caption{Method of intermediate steps in shock wave extraction: Events are re-encoded into polar coordinate form firstly. Subsequently, for any propagation angle, construct a $d-t$ graph. Furthermore, the adaptive ROI extraction algorithm is utilized.}
\end{figure}

\subsubsection{Polar coordinate encoding and d-t diagram}

The polar coordinate modeling method demonstrates clear advantages in measurement tasks involving circular targets\cite{ref31}\cite{ref32}. This is primarily due to the ability of polar coordinates to accurately capture the radial and tangential motion parameters of these targets. Therefore, polar coordinate modeling is also well-suited for analyzing shock waves.

Shock waves propagate outward from the center of the blast following the explosion. Events are triggered at the shock wave front due to a sharp change in air density, which affects the refractive index. Consequently, the blast center serves as a crucial reference point during shock wave propagation, as illustrated by the following two aspects.

1. The distance between the shock wave front and the blast center is critical for accurately measuring motion parameters, such as the shock wave radius.

2. To estimate the asymmetry of shock wave propagation, measurements must be taken at various propagation angles. The shock wave front in a given direction forms a vector with the blast center, and the angle between this vector and the positive $x$-axis serves as a key reference feature.

Therefore, all events are transformed into polar coordinate encoding with the blast center as the pole:
	\begin{align}
              \left\{
              \begin{aligned}
                   d&=\parallel(x-x_{b},y-y_{b})\parallel\\
                   \alpha&=\cos^{-1}{(\frac{x-x_{b}}{d})}
              \end{aligned}
              \right.
        \end{align} 
where $d$ is the distance between the shock wave front and the blast center, $\alpha$ is the angle mentioned above, $x_{b}$ and $y_{b}$ are coordinates of the blast center. Through the transformation described above, the event's universal encoding format $e_s(x_s,y_s,p_s,t_s)$ is converted into $e_s(d_s,\alpha_s,p_s,t_s)$. All events are then partitioned into several segments $E_1,E_2,...E_l...$ based on the angle $\alpha_s$. This partitioning is intended to assess shock waves at various propagation angles. 

Upon the arrival of the blast, events are primarily triggered by the blast firelight, blast products, noise, and shock waves. The blast firelight, though brief, covers a wide field of view, while blast products have a longer duration but a limited distribution, quickly surpassed by shock waves. Noise is irregular and sparse. Shock waves arrive after the firelight and exceed the distribution area of the blast products, exhibiting significant continuity. Using polar coordinate encoding, distinct features of events triggered by different objects are emphasized for any propagation angle $\alpha_l$. The two-dimensional distance-over-time histogram then projects the events of $E_l$ on their $d$ and $t$ values. As shown in Fig. 3, events triggered by the blast firelight cluster within a short time interval but cover a large distance, forming a nearly vertical line. In contrast, events from blast products are densely distributed in regions with small distance spans and large time spans, while noise remains sparse. Events triggered by shock waves create a distinct 'edge' to the right of those from the blast firelight, surrounding the events from the blast products. This polar coordinate encoding unlocks the prominent features of the shock wave front in the event camera.

\subsubsection{Adaptive ROI extraction algorithm}

 Directly processing all events is inefficient. Therefore, our goal is to select candidate events for each angle to facilitate efficient extraction. To swiftly select region of interest (ROI) and narrow the scope of event extraction, two reference points $(t_1,d_1)$, $(t_2,d_2)$ are selected. These points represent the starting and ending locations of the shock wave front over a relatively long time interval. They are selected using two fan-shaped annular monitoring windows, based on event statistical principles. Firstly, we leverage the significant temporal and spatial differences in the triggering events of shock waves, firelight, and smoke to eliminate firelight and smoke. Next, we select the direction with the highest event statistics to ensure the robustness of the initial reference point extraction. Finally, we choose the temporal and spatial median events within the window as the initial reference points. Subsequently, the ROI is identified according to:
	\begin{align}
              \left\{
              \begin{array}{l}
                   \displaystyle d \ge d_{1}-q'+\epsilon+\frac{(t-t_1)(d_2-d_1)}{t_2-t_1}\\
                   \displaystyle d\le d_{1}+\epsilon+\frac{(t-t_1)(d_2-d_1)}{t_2-t_1}\\
                   \displaystyle t_{1}\le t\le t_{2}, d_{1}\le d\le d_{2}
              \end{array}
              \right.
        \end{align} 
where $q'$ is the search radius determined through an expansion approach that leverages the feature of an increasing number of events in the ROI. When the search radius reaches a certain threshold $q'$ and the number of events no longer increases significantly, the search radius value is returned. $\epsilon$ is the iterative correction value. For example, if the final extracted shock wave event is positioned too close to the upper bound of the ROI, the value of $\epsilon$ for the next angle $\alpha _{l+1}$ is adaptively updated as follows:
	\begin{align}
              \left\{
              \begin{array}{l}
                   \displaystyle \epsilon=\frac{T-min(S^{'},\ell)}{\cos{(\tan^{^{-1}}{(\displaystyle\frac{d_2-d_1}{t_2-t_1})})}},\quad \\if \quad
                    min(S^{'},\ell)\le T \quad and\\
                   \displaystyle  T \le q'\cos{(\tan{^{-1}}{(\frac{d_2-d_1}{t_2-t_1})})}-max(S^{'},\ell)
              \end{array}
              \right.
        \end{align} 
where $T$ is the tolerance threshold, $\ell$ is the line composed of points $(t_1,d_1)$ and $(t_2,d_2)$, $S^{'}$ is the final shock wave extraction result introduced in section $III.C$, $min(S^{'},\ell)$ denotes the minimum value from events in set $S^{'}$ to line $\ell$, as well as $max(S^{'},\ell)$. The trigonometric function in Eq. (7) is primarily used to perform geometric transformations between the hypotenuse and adjacent sides, enabling the conversion between pixel coordinates and Euclidean distance. All events' $(d,t)$ satisfied the above equation are chosen as candidates. Although shock waves propagate differently in various directions, differences between adjacent angles fall within a small range. Therefore, our method can adaptively extract ROI for multiple angles, demonstrating its versatility.

\subsection{Slope-iterative-based shock wave extraction algorithm}

In this subsection, we aim to select shock wave front events from all candidate events within the ROI. The velocity of shock waves changes gradually during propagation, rather than exhibiting abrupt attenuation. Consequently, our search algorithm is based on slope iteration. The entire process is shown in the Fig. 4.
\begin{figure}[h]
  \centering
  \includegraphics[width=0.48\textwidth,height=!]{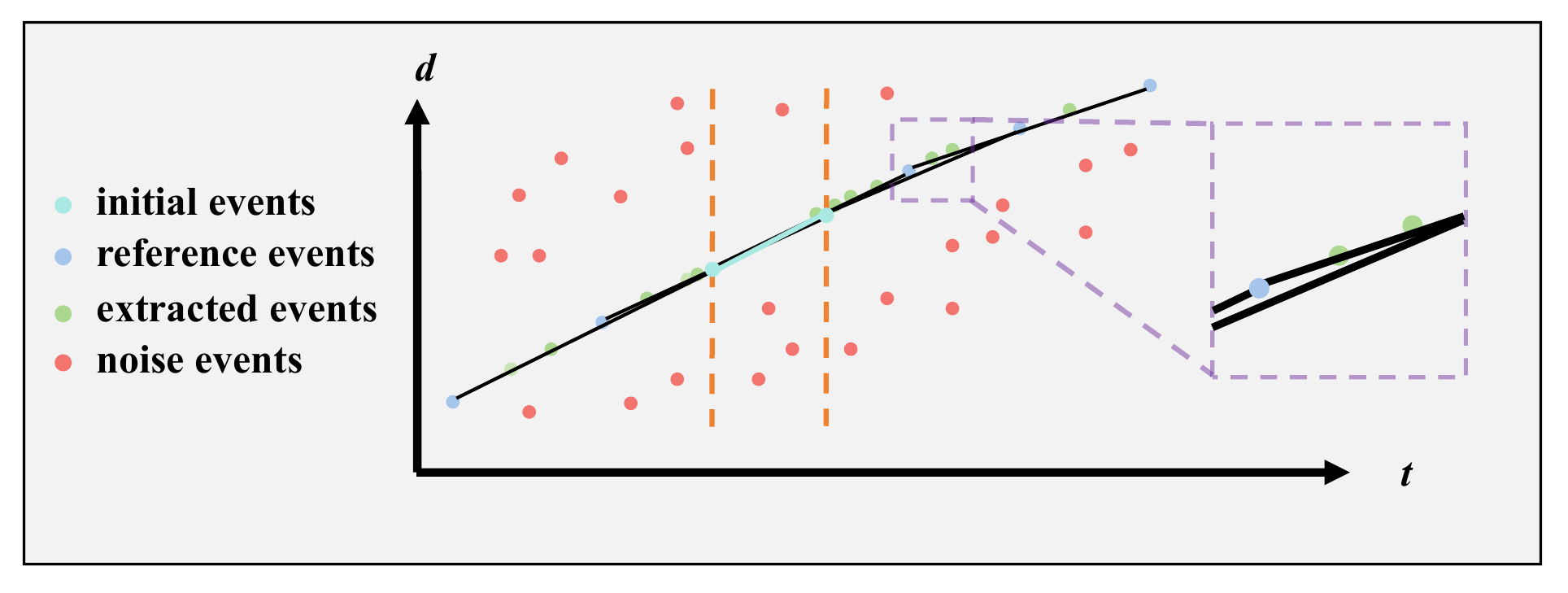}
  \caption{Method of the slope-iterative extraction algorithm. As the search process advances, the slope is updated iteratively in small increments.}
\end{figure}

First, the density feature is utilized to identify the two most likely shock wave events in the array, as the event density triggered by the array is significantly higher than that of the noise. The density $\sigma _s$ of an event $e_s$ is defined as follows:
	\begin{eqnarray}
              \sigma _s=\sum_{t=t_s-q}^{t_s+q}\sum_{d=d_s-q}^{d_s+q}\mid p(d,t) \mid
        \end{eqnarray}
where $q$ is the search radius, $p(d,t)$ is the polarity of the event with coordinates $(d,t)$. Two events, which pass density checks and are separated by specific intervals, are established as initial points for calculating the reference slope $k_\tau$. Afterwards, all candidate events are divided into three sets based on their $t$: front($S_1$), middle($S_2$), and back($S_3$):
	\begin{eqnarray}
              \left\{
              \begin{array}{l}
                   S_1\leftarrow e_s,\quad if \quad t_s\le t'\\
                   S_2\leftarrow e_s,\quad if \quad t' \le t_s\le t''\\
                   S_3\leftarrow e_s,\quad if \quad t''\le t_s\\
                   t'=\min(t_m,t_n),\quad t''=\max(t_m,t_n)
              \end{array}
              \right.
        \end{eqnarray} 
where $t_m$ is the timestamp of the event with the highest density, $t_n$ is the timestamp of the event with the second highest density. Each event $e_s$ is used to calculate its slope feature $k_s$ according to:
	\begin{eqnarray}
            k_s=\frac{d_s-d_\varsigma}{t_s-t_\varsigma}
        \end{eqnarray}
where $\varsigma$ is the index of the reference event $e_\varsigma$. For each event in the middle set $S_2$, its reference event is determined as the farther one between $e'(d',t')$ or $e''(d'',t'')$. In the front set $S_1$, events are traversed from back to front according to timestamps, with the initial reference event $e_\varsigma$ being $e''(d'',t'')$. As the traversal progresses, the slope $k_s$ of an event $e_s$ is compared with its reference slope $k_\tau$:
	\begin{eqnarray}
            S^{'}\leftarrow e_s,\quad if \mid k_s-k_\tau \mid \le \rho
        \end{eqnarray} 
where $\rho$ is the threshold. Once the distance between the current event $e_s(d_s,t_s)$ and the first benchmark event $e_\iota(d_\iota,t_\iota)$ exceeds the threshold, the reference slope $k_\tau$ is updated to $(d_s-d_\iota)/(t_s-t_\iota)$. The reference event is then updated to $e_\iota(d_\iota,t_\iota)$, and the benchmark event is set to $e_s(d_s,t_s)$. The same operation is applied to the events of $S_3$ until all events have been traversed.
    \begin{algorithm}[t]
    \renewcommand{\algorithmicrequire}{\textbf{Input:}}
	\renewcommand{\algorithmicensure}{\textbf{Output:}}
	\caption{Shock wave events search algorithm}
    \label{power}
    \begin{algorithmic}[1]
        \REQUIRE Event set $S_1$, initial reference slope $k_\tau$, initial reference event $e_\varsigma$, initial benchmark event $e_\iota$, tolerate threshold $\rho$;
	    \ENSURE Shock wave events set $S^{'}$;
        \FOR {$s=card(S_1),card(S_1)-1,\cdots,2,1$}
            \STATE Count $k_s$ with Eq.(10);
        \IF {$\mid k_s-k_\tau \mid \le \rho$}
            \STATE Check distance between $e_s$ and $e_\iota$
        \IF {Passed}
            \STATE $k_\tau=\displaystyle \frac{d_s-d_\iota}{t_s-t_\iota}$; \STATE $\varsigma=\iota$; \STATE $\iota=s$;
        \ENDIF
            \STATE $S^{'}\leftarrow e_s$;
        \ENDIF 
        \ENDFOR
        \STATE \textbf{return} $S^{'}$.
    \end{algorithmic}
\end{algorithm}
Finally, all events triggered by the shock wave front are included in set $S^{'}$, which represents the result of the shock wave extraction. The iterative search process for set $S_1$ is detailed in Algorithm 1. Importantly, we have preserved the original data information of each event throughout the extraction process. As a result, our shock wave extraction achieves a temporal resolution at the microsecond level.

\subsection{Event-based shock waves measurement and 3D motion field reconstruction}

Through the previous steps, we extracted the shock wave front event set $S^{'}_l$ for each propagation angle $\alpha_l$. By combining this with intrinsic parameters, the motion parameters of shock waves can be measured. In this subsection, the geometric model relating the radius of the shock wave to the events is derived. The model allows for shock wave radius estimation without imposing excessive limitations on shooting conditions. Using the extrinsic parameters and other view data, the 3D motion field of the shock waves is reconstructed. Additionally, the equivalent of the charge is reversed.
\begin{figure}[h]
  \centering
  \includegraphics[width=0.45\textwidth,height=!]{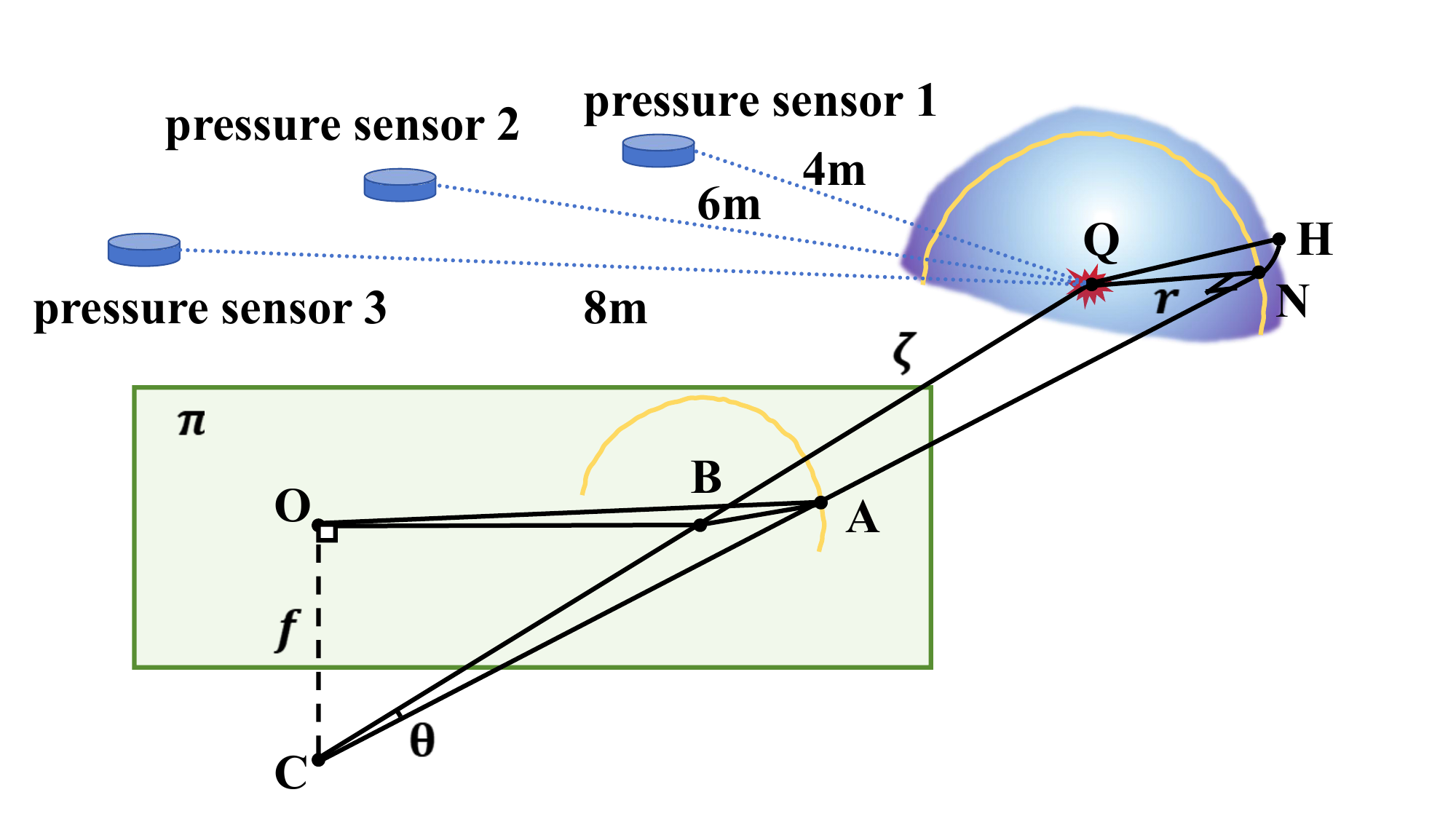}
  \caption{Geometric model relating the radius of the shock wave to the events. By employing this model, the transformation from event coordinates to 3D radius can be achieved.}
\end{figure}

\subsubsection{Multi-view and propagation angles motion measurement}

 In this part, a geometric model relating the radius of the shock wave to the events is derived for measurement. As shown in Fig. 5, $C$ represents the optical center of the event camera, $\pi$ denotes the image plane, $O$ is the principle point, $A$ is the image point of an event on the shock wave front, $B$ is the image point of blast center and $Q$ is the blast center. $H$ is any point along the straight line that is perpendicular to $\overline{CQ}$ and passes through the explosion center $Q$, intersecting with the shock wave front. Since the actual imaging observation point of the shock wave is defined by the set of tangent points (such as $N$) , formed by the intersections of light rays and cameras' principle point with the shock wave front, the observation event at point $A$ of the shock wave corresponds to point $N$ rather than point $H$. The length of $\overline{OC}$ represents the focal length $f$ and the coordinates of $O$ correspond to the principal point. According to our extraction results, $\overline{OB}$ and $\overline{OA}$ are known. Since $\overline{CO}\perp \pi$, the length of $\overline{AB}$, $\overline{CB}$ and $\overline{CA}$ can be calculated with:
	\begin{align}
              \left\{
              \begin{aligned}
                   \overline{CA}&=\parallel (x_o-x_a,y_o-y_a,f) \parallel\\
                   \overline{CB}&=\parallel (x_o-x_b,y_o-y_b,f) \parallel\\
                   \overline{AB}&=\parallel (x_a-x_b,y_a-y_b) \parallel
              \end{aligned}
              \right.
        \end{align} 
where $x_o,y_o$ are the image coordinates of $O$, and similarly for $A$, $B$. Furthermore, the angle $\theta$ is calculated based on Eq.(12) and the Cosine theorem: 
	\begin{eqnarray}
            \theta=G_1(\mathcal{X})=\cos^{-1}{(\frac{\overline{CA}^2+\overline{CB}^2-\overline{AB}^2}{2\overline{CB}\cdot\overline{CA}})}
        \end{eqnarray}
Here, the function $G_1$ is used to represent the equation for simplicity in the following text. The $\mathcal{X}$ represents representing variables $x_o$, $y_o$, $x_a$, $y_a$, $x_b$, $y_b$, and $f$. In fact, the length of $\overline{QN}$ represents the radius of shock wave. $\overline{QC}$ is the distance $\zeta$ between the blast center and the optical center. Since the observed shock wave front is formed by the tangent of light to the shock wave hemisphere in space, we have $\overline{QN}\perp \overline{NC}$. Therefore, combined with Eq.(12) and Eq.(13), the microsecond-level radius $r$ can be estimated according to:
	\begin{eqnarray}
            r=G_2(\zeta,\theta)=\zeta\sin{\theta}
        \end{eqnarray}
Here, the function $G_2$ is used to represent the equation for simplicity in the following text. Through the above solution, for each event $e_s$, there is a corresponding radius $r_s$. The polynomial $\Psi$ is then used to fit the scatter data of the radius over time. The velocity $v$ of the shock wave can be estimated by taking the derivative of the radius with respect to time. This approach allows us to avoid the significant errors associated with estimating speed using time intervals.

The shock wave motion parameters are measured for each propagation angle using the method described above, yielding the results for a single viewing angle. Furthermore, these measurements were extended to multiple event cameras, ultimately achieving the measurement of shock wave motion parameters from various propagation angles.

\subsubsection{3D reconstruction of shock wave motion field}

Based on our previous method, we finally obtain fitting polynomials $\Psi_{\lambda\alpha}$ for multiple views and propagation angles. Here, $\lambda(=1,2,...)$ represents the index of views, while $\alpha$ denotes the propagation angle in each view. To visualize the shock wave motion field more vividly, projections from various views and 3D reconstructions are conducted. Using the fitting polynomials $\Psi_{\lambda\alpha}$, the same timestamps are selected to calculate the radius for different angles $\alpha$. Subsequently, equations are established based on Eq.(12), Eq.(13) and Eq.(14) to estimate the image points:
	\begin{eqnarray}
              \left\{
              \begin{array}{l}
                   \overline{AB}^2+2\overline{CB}\cdot\overline{CA}\cos{\theta}=\overline{CB}^2+\overline{CA}^2
                   \\
                   x_{\lambda\alpha}-x_{\lambda b}=\parallel (x_{\lambda\alpha}-x_{\lambda b},y_{\lambda\alpha}-y_{\lambda b})\parallel \cos{\alpha}
              \end{array}
              \right.
        \end{eqnarray} 
where $(x_{\lambda b},y_{\lambda b})$ denote the coordinates of the blast center as detected by event camera $\lambda$, $(x_{\lambda o},y_{\lambda o})$ represents the principal point of event camera $\lambda$. Our objective is to determine the coordinates $(x_{\lambda\alpha},y_{\lambda\alpha})$. 

Furthermore, the 3D spatial points $(X_{\alpha},Y_{\alpha},Z_{\alpha})$ corresponding to the shock wave front points $(x_{\lambda\alpha},y_{\lambda\alpha})$ are estimated to facilitate reconstruction. For each event camera, the projection matrix $\bm{\Gamma_\lambda}$ can be synthesized according to its intrinsic and extrinsic parameters.
	\begin{eqnarray}
            \bm{\Gamma_\lambda}=
           \left [
            \begin{array}{cc}
                \gamma_{11}\quad \gamma_{12}\quad \gamma_{13}\quad \gamma_{14}\\
                \gamma_{21}\quad \gamma_{22}\quad \gamma_{23}\quad \gamma_{24}\\
                \gamma_{31}\quad \gamma_{32}\quad \gamma_{33}\quad \gamma_{34} 
            \end{array}
           \right ]
        \end{eqnarray}
Subsequently, the equation system for $(X_{\alpha},Y_{\alpha},Z_{\alpha})$ is established:
	\begin{eqnarray}
              \begin{array}{l}
                   {\parallel (X_{\alpha}-X_{\lambda C},Y_{\alpha}-Y_{\lambda C},Z_{\alpha}-Z_{\lambda C}) \parallel}^2=\zeta_{\lambda}^2-r_{\lambda\alpha}^2
                   \\\\
                   \parallel (X_{\alpha}-X_{B},Y_{\alpha}-Y_{B},Z_{\alpha}-Z_{B}) \parallel=r_{\lambda\alpha}\\\\
                   
                   {\bm{\Gamma_\lambda}}_{1,:}\left [
            \begin{array}{cc}
                X_{\alpha} \\ Y_{\alpha}\\Z_{\alpha}\\ 1
            \end{array}
            \right ]=x_{\lambda\alpha}({\bm{\Gamma_\lambda}}_{3,:}\left [
            \begin{array}{cc}
                X_{\alpha}\\ Y_{\alpha}\\ Z_{\alpha}\\ 1
            \end{array}
            \right ])\\
                                \\{\bm{\Gamma_\lambda}}_{2,:}\left [
            \begin{array}{cc}
                X_{\alpha}\\ Y_{\alpha}\\ Z_{\alpha}\\ 1
            \end{array}
            \right ]=y_{\lambda\alpha}({\bm{\Gamma_\lambda}}_{3,:}\left [
            \begin{array}{cc}
                X_{\alpha}\\ Y_{\alpha}\\ Z_{\alpha}\\ 1
            \end{array}
            \right ])
              \end{array}
        \end{eqnarray} 
where $(X_{\lambda C},Y_{\lambda C},Z_{\lambda C})$ represents the 3D coordinates of the optical center of event camera $\lambda$, $(X_{B},Y_{B},Z_{B})$ denotes the 3D coordinates of the blast center. Upon solving Eq.(17) using the Gaussian elimination method, the 3D points of the shock wave front can be visualized. For each view, others 3D points of shock wave front are projected with $\bm{\Gamma_\lambda}$.
	\begin{eqnarray}
            \left [
            \begin{array}{cc}
                x\quad y\quad 1
            \end{array}
            \right ]^T=\bm{\Gamma_\lambda}
            \left [
            \begin{array}{cc}
                X\quad Y\quad Z\quad 1
            \end{array}
            \right ]^T.
        \end{eqnarray}

\begin{figure*}
  \centering
  \begin{subfigure}{0.38\linewidth}
  \centering
  \includegraphics[width=\textwidth,height=0.5\textwidth]{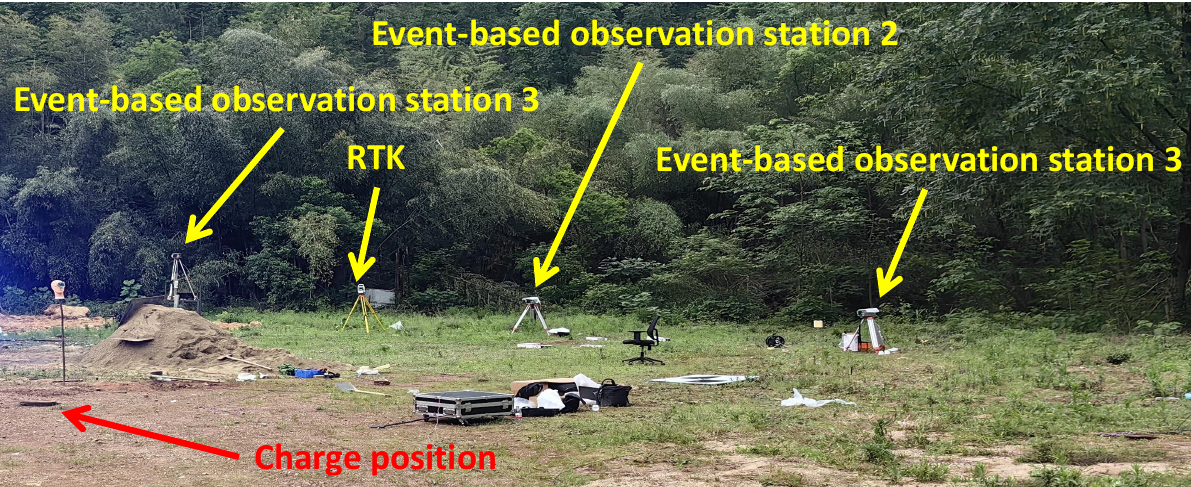}
  \caption{Experimental field and layout}
  \end{subfigure}
  \begin{subfigure}{0.19\linewidth}
    \centering
  \includegraphics[width=\textwidth,height=\textwidth]{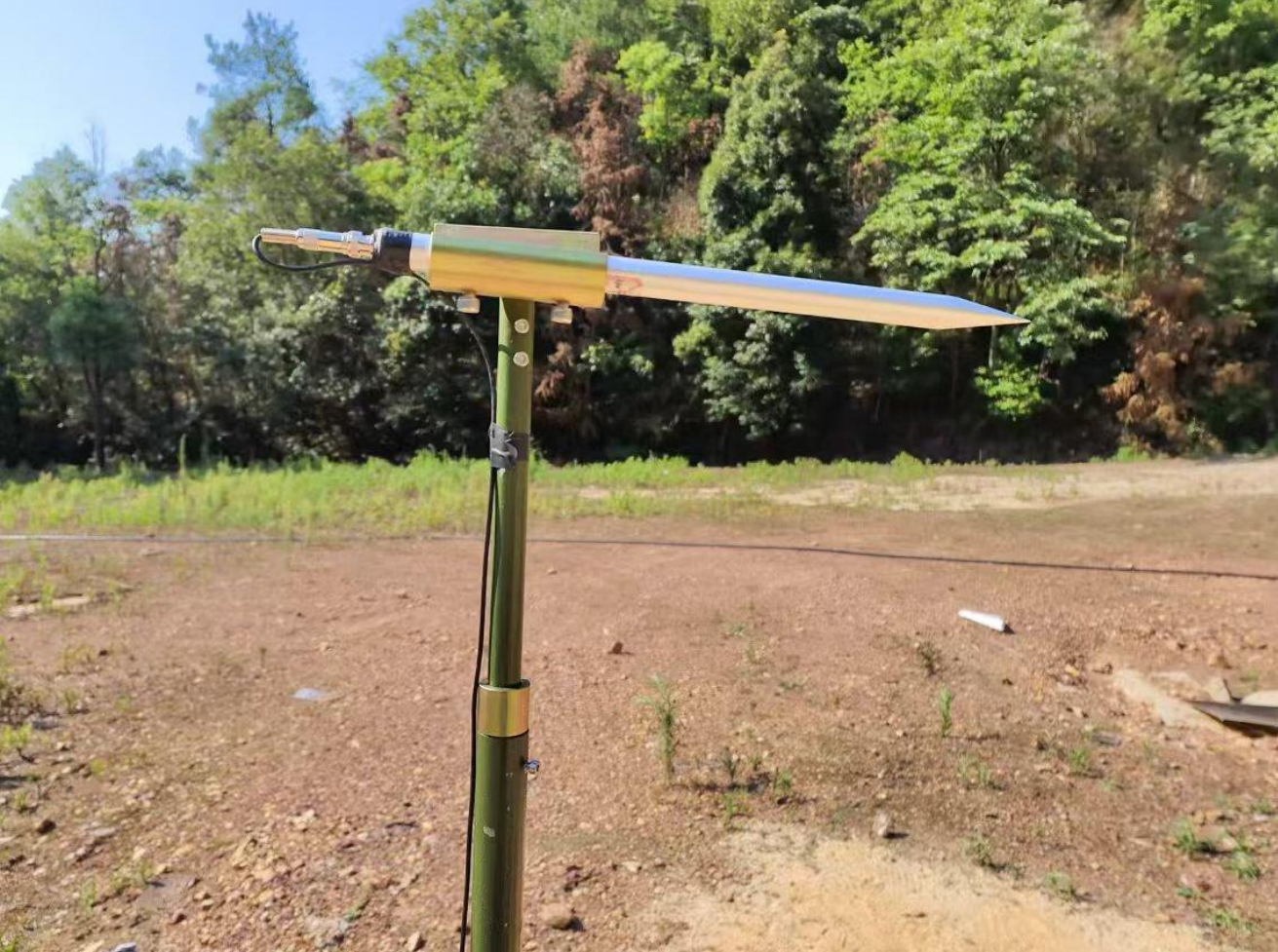}
  \caption{Free field pressure sensor}
  \end{subfigure}  
  \begin{subfigure}{0.19\linewidth}
    \centering
  \includegraphics[width=\textwidth,height=\textwidth]{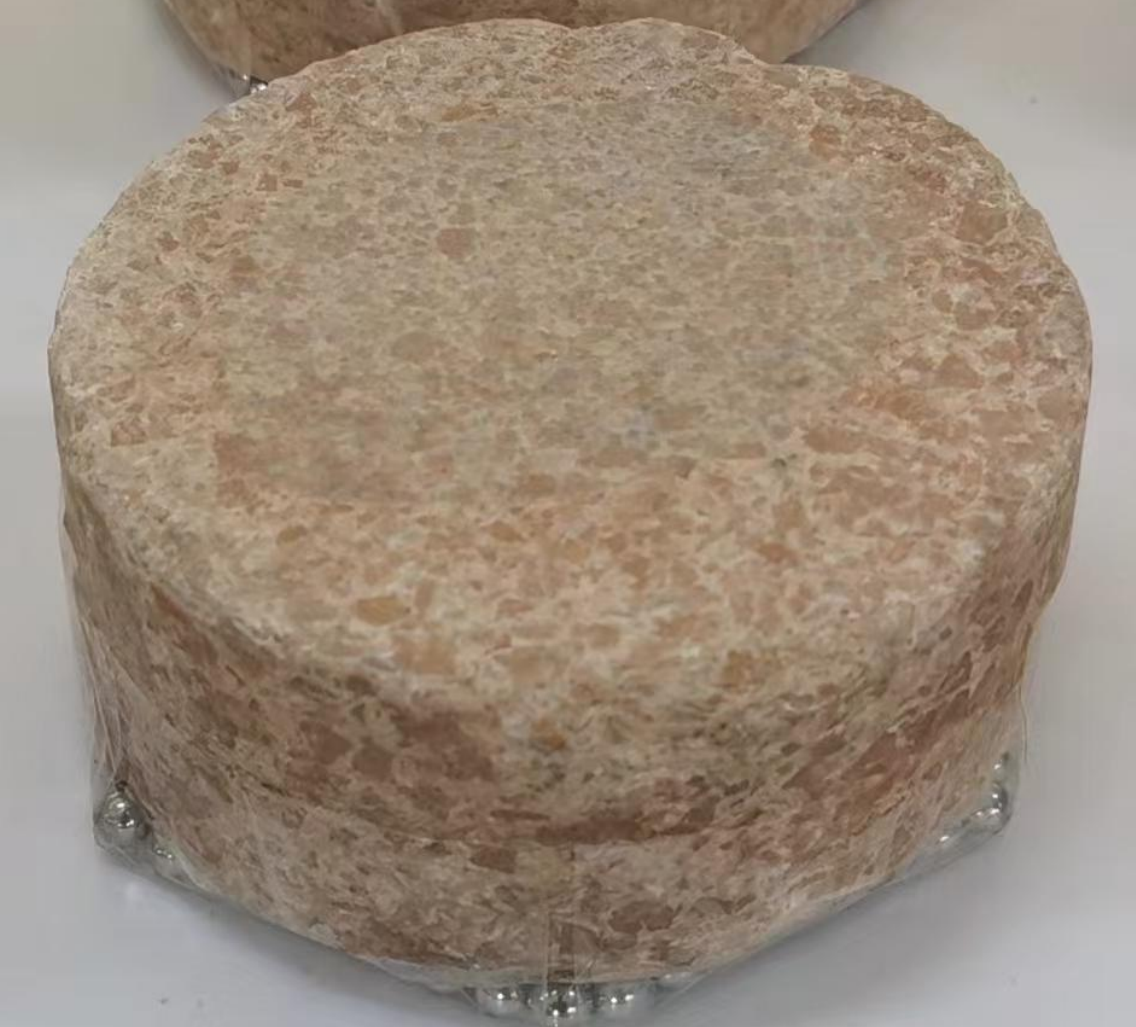}
  \caption{TNT charge}
  \end{subfigure}  
  \begin{subfigure}{0.19\linewidth}
    \centering
  \includegraphics[width=\textwidth,height=\textwidth]{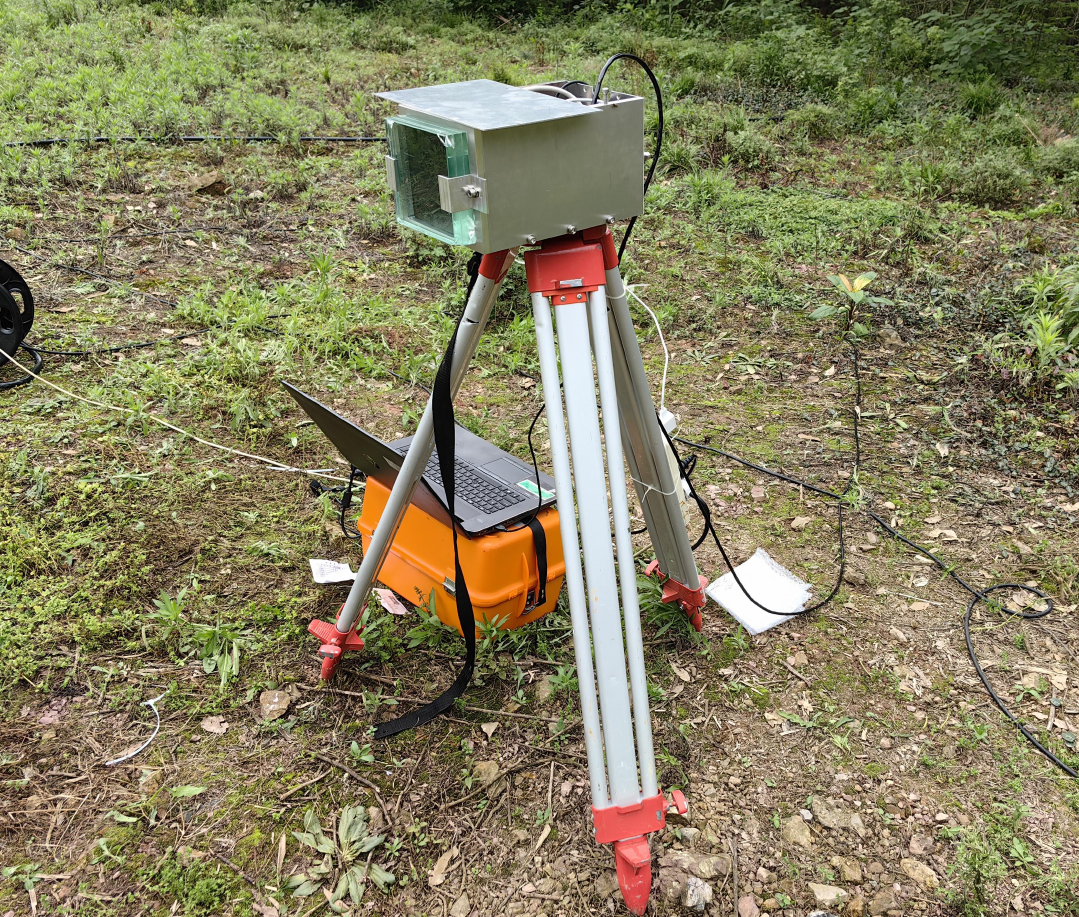}
  \caption{Our observation station}
  \end{subfigure}  
  \caption{Experimental field and equipment}
\end{figure*}

\begin{table*}
  \centering
    \caption{Calibration results of our multiple-event camera observation system}
  \begin{tabular}{c c c c}
  \hline
  Event Cameras & Left Camera & Middle Camera & Right Camera \\
  \hline
  Focal Length (pixels) & $856.54$ & $855.27$ & $852.26$ \\
  Principle Point (pixels) & $(632.04, 355.46)$ & $(631.90, 364.76)$ & $(631.09, 342.95)$ \\
 Optical Center $(\mathrm{m})$ & $(-20.37, 1.13, -2.66)$ & $(-9.97, 0.70, -2.89)$ & $(-1.78,  0.42, -0.44)$ \\
 Euler Angle $(\mathrm{rad})$ & $(-0.2850, -0.0259, -1.591)$ & $(0.276, 0.009, -1.596)$ & $(0.475, -0.009, -1.666)$ \\
 Translation Vector $(\mathrm{m})$ & $(19.93, -2.24, 4.60)$ & $(9.37, -2.89, -3.47)$ & $(1.40, -0.31, -1.22)$ \\
 Reprojection Error (pixels) & $0.35$ & $0.38$ & $0.34$\\
  \hline
  \end{tabular}
\end{table*}

Finally, the irregular surface with a known explosion center is used to fit all 3D points, enabling detailed visualization of the shock wave propagation.

\subsubsection{Charge equivalent reversion}

Once the velocity of the shock wave front is measured, the overpressure of the shock wave can be estimated using the Rankine-Hugoniot condition\cite{ref33} from:
	\begin{eqnarray}
            P=\frac{2\eta}{\eta+1}(M^2-1)P_0
        \end{eqnarray}
where $P$ is the overpressure, $\eta$ ($\approx 1.4$) is the specific heat ratio, $M$ is the Mach number of velocity, and $P_0$ ($\approx 1.01325kPa$) is the ambient atmospheric pressure. Therefore, the equivalent of the charge can be derived using the empirical formula \cite{ref34} for shock wave overpressure:
	\begin{eqnarray}
            P=\frac{0.108}{\delta}+\frac{0.114}{\delta^2}+\frac{1.772}{\delta^3}
        \end{eqnarray}
where $\delta=r/\sqrt[3]{W}$ is the scaled distance which normalized the blast generally. Then, the equivalent $W$ of the charge is reversed.

\section{Experiments}

In our experiments, multiple event cameras housed in protective boxes are positioned at appropriate intersection angles at the blast test site. Synchronous acquisition is achieved using a controller that emits pulse signals. The middle event camera serves as the master, while the other two are configured as slaves. The synchronization accuracy of the synchronization device is less than $10\upmu\mathrm{s}$, verified using a LED marker with a known flicker rate. The $\mathrm{4mm}$ lenses are chosen to ensure covering a large field of view. The blast charge consists of TNT with an equivalent mass of $\mathrm{600g}$, positioned approximately $\mathrm{1.5m}$ above the ground. Free field pressure sensors (IEPE KD2002L) are deployed in various directions at distances of 4, 6, and 8 meters from the blast center. RTK positioning is employed to assist in calibration and to establish a world coordinate system. The entire experimental setup is illustrated in Fig. 6.

\subsection{LED-based event cameras system calibration}

Our multiple-event camera observation system is calibrated using the method introduced in section III.A. First, the 2D coordinates of the LED marker are extracted, as illustrated in Fig. 7. By combining these with the corresponding 3D world coordinates obtained from RTK, the calibration results of our system are presented in TABLE I.
\begin{figure}[h]
  \centering
  \includegraphics[width=0.45\textwidth,height=!]{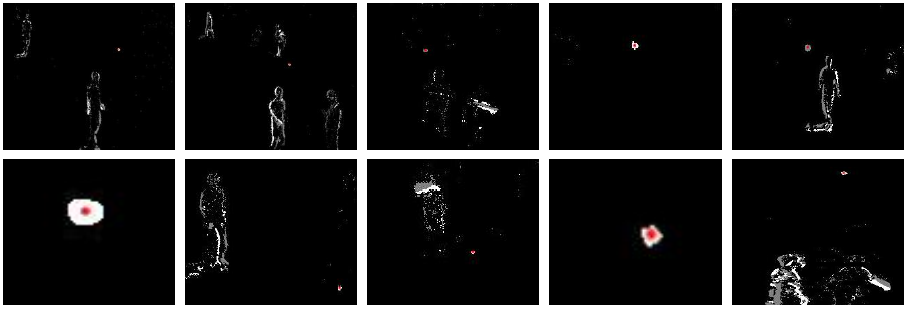}
  \caption{Extraction results of the LED marker. The red dot represents the extraction results.}
\end{figure}

\subsection{Microsecond-level extraction of shock waves}

In our experiment, short event stamps recording the shock wave propagation from each event camera are extracted. All events are then encoded in polar coordinates to construct the $d-t$ diagrams for each propagation angle, as detailed in sections III.B.1 and III.B.2. The $d-t$ diagrams of each view for $\alpha=55^{\circ}$ and the ROI extraction results from event cameras are shown in Fig. 8 (left). Subsequently, events triggered by the shock wave front are extracted using the search algorithm described in section III.C. The results of the slope iteration and event extraction are presented in Fig. 8 (right).
\begin{figure}[h]
  \centering
  \includegraphics[width=0.45\textwidth,height=!]{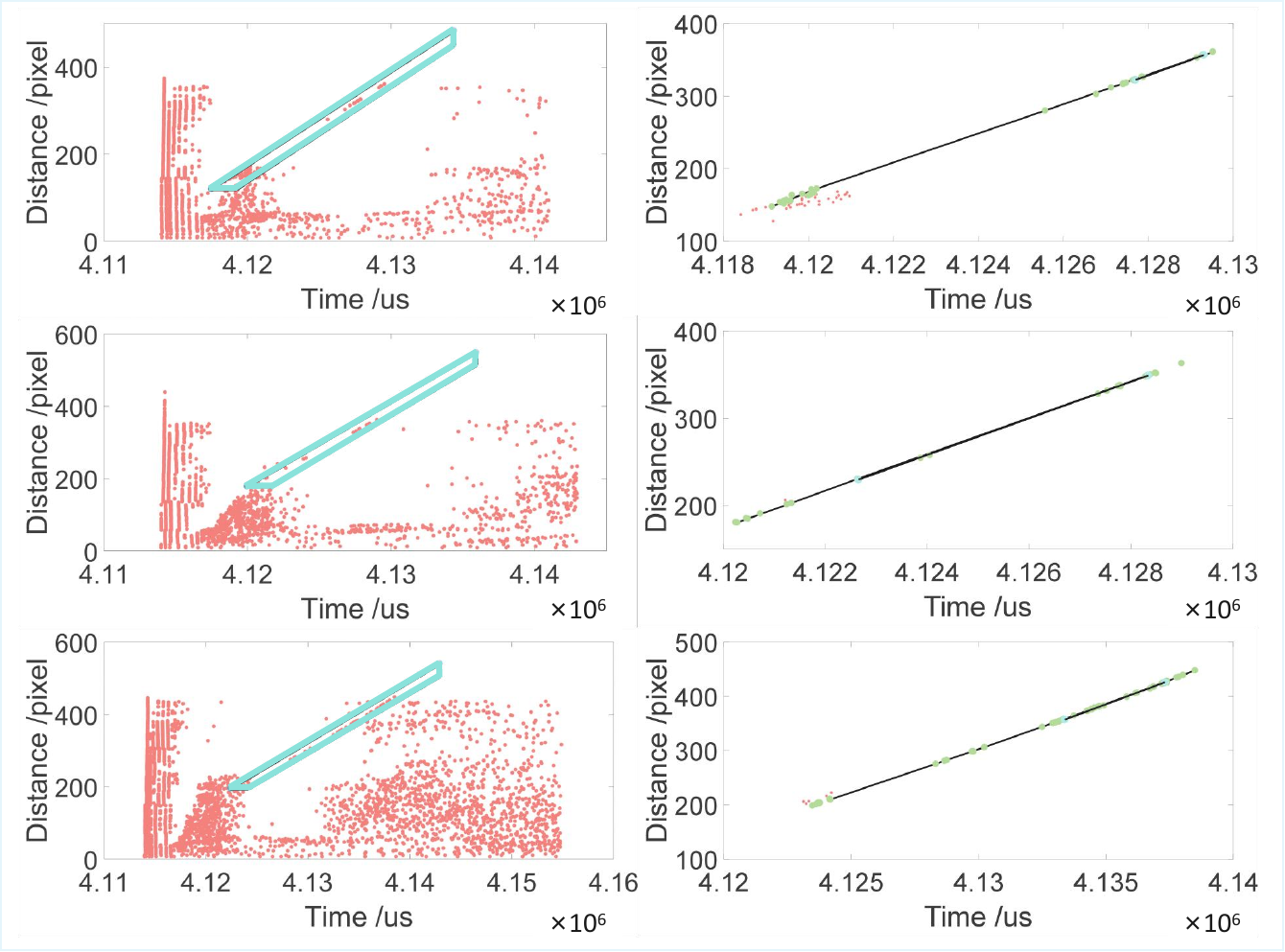}
  \caption{Intermediate processing results of our extraction algorithm. Each red dot represent an event, while the yellow dots represent extraction events. The black line segments (right) denotes the reference line segment during the iteration process.}
\end{figure}

The complete extraction results for various propagation angles are obtained after processing all the data. To facilitate the demonstration of our extraction results, events are accumulated into event frames, with the results displayed on the event frame images. Examples of shock wave extraction sequences are shown in Fig. 9. It is important to note that our extraction operates at the microsecond level events, while the cumulative frame building time for the displayed images is $200\upmu\mathrm{s}$. This results in a "thickening" effect of the shock wave front in the event frame images, which indirectly highlights the superiority of our method. Specifically, our extraction directly processes events without constructing images, ensuring accurate shock wave front extraction, and achieves a time resolution of $1\upmu\mathrm{s}$. To our knowledge, this is currently the most sensitive method available.
\begin{figure}[h]
  \centering
  \begin{subfigure}{\linewidth}
  \centering
  \includegraphics[width=\textwidth,height=!]{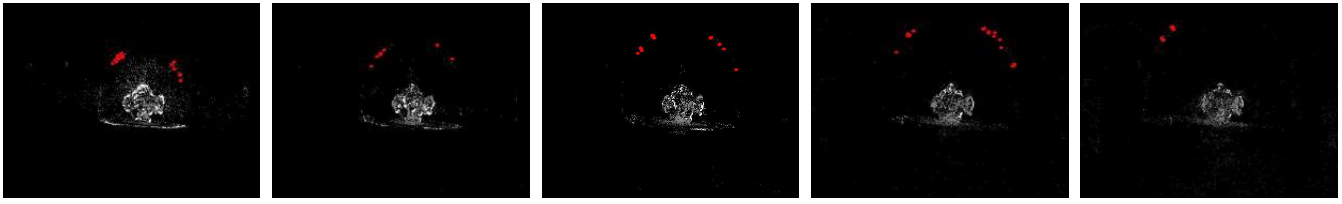}
  \caption{Extraction results of slave event camera (left)}
  \end{subfigure}
  \qquad
  \begin{subfigure}{\linewidth}
  \centering
  \includegraphics[width=\textwidth,height=!]{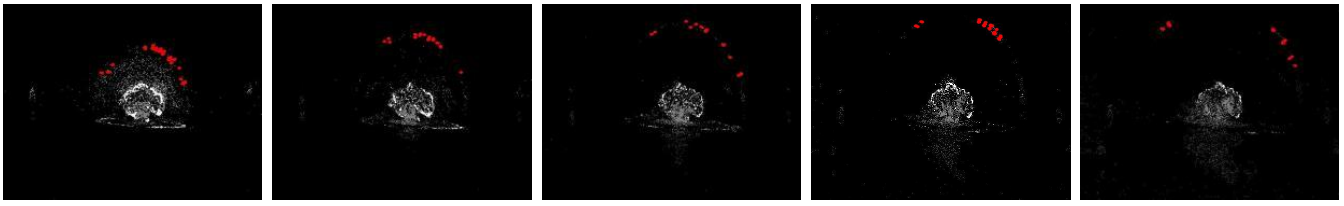}
  \caption{Extraction results of master event camera (middle)}
  \end{subfigure}
  \qquad
  \begin{subfigure}{\linewidth}
  \centering
  \includegraphics[width=\textwidth,height=!]{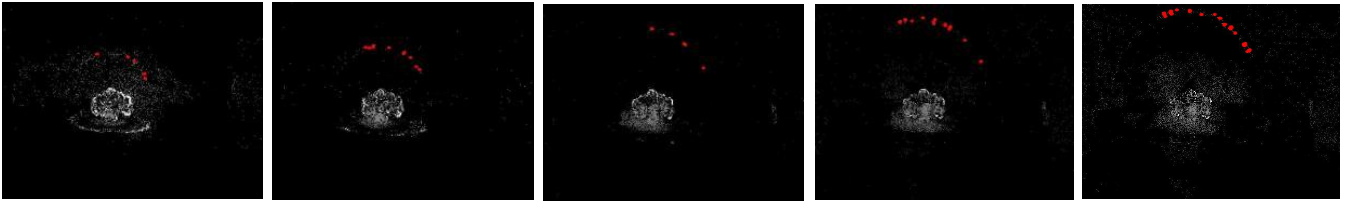}
  \caption{Extraction results of slave event camera (right)}
  \end{subfigure}
  \caption{Extraction results of shock wave fronts. The red points are our micro-level second extraction results.}
\end{figure}

\begin{figure}[h]
  \centering
  \begin{subfigure}{\linewidth}
  \centering
  \includegraphics[width=0.9\textwidth,height=!]{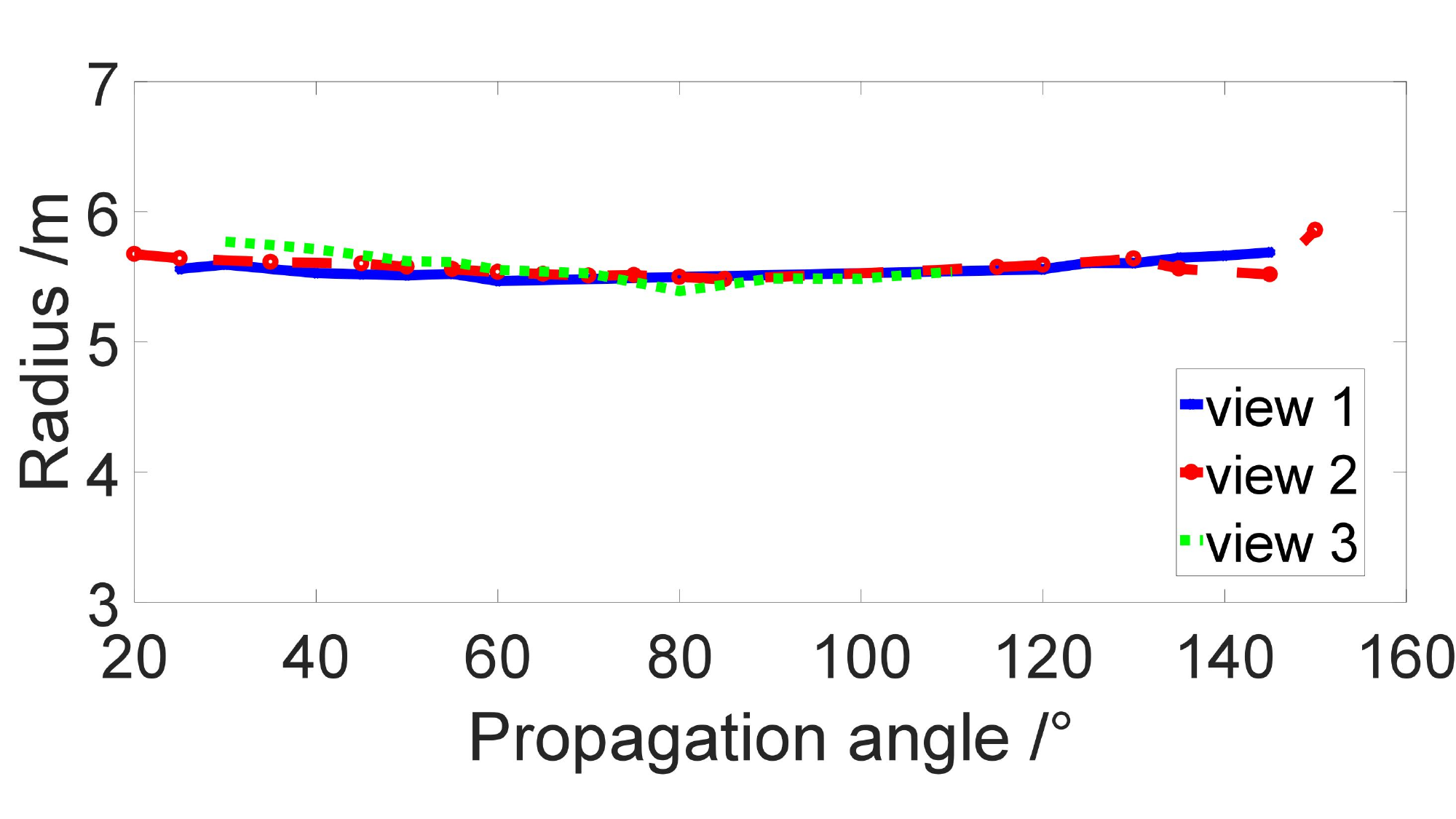}
  \end{subfigure}
  \qquad
  \begin{subfigure}{\linewidth}
  \centering
  \includegraphics[width=0.9\textwidth,height=!]{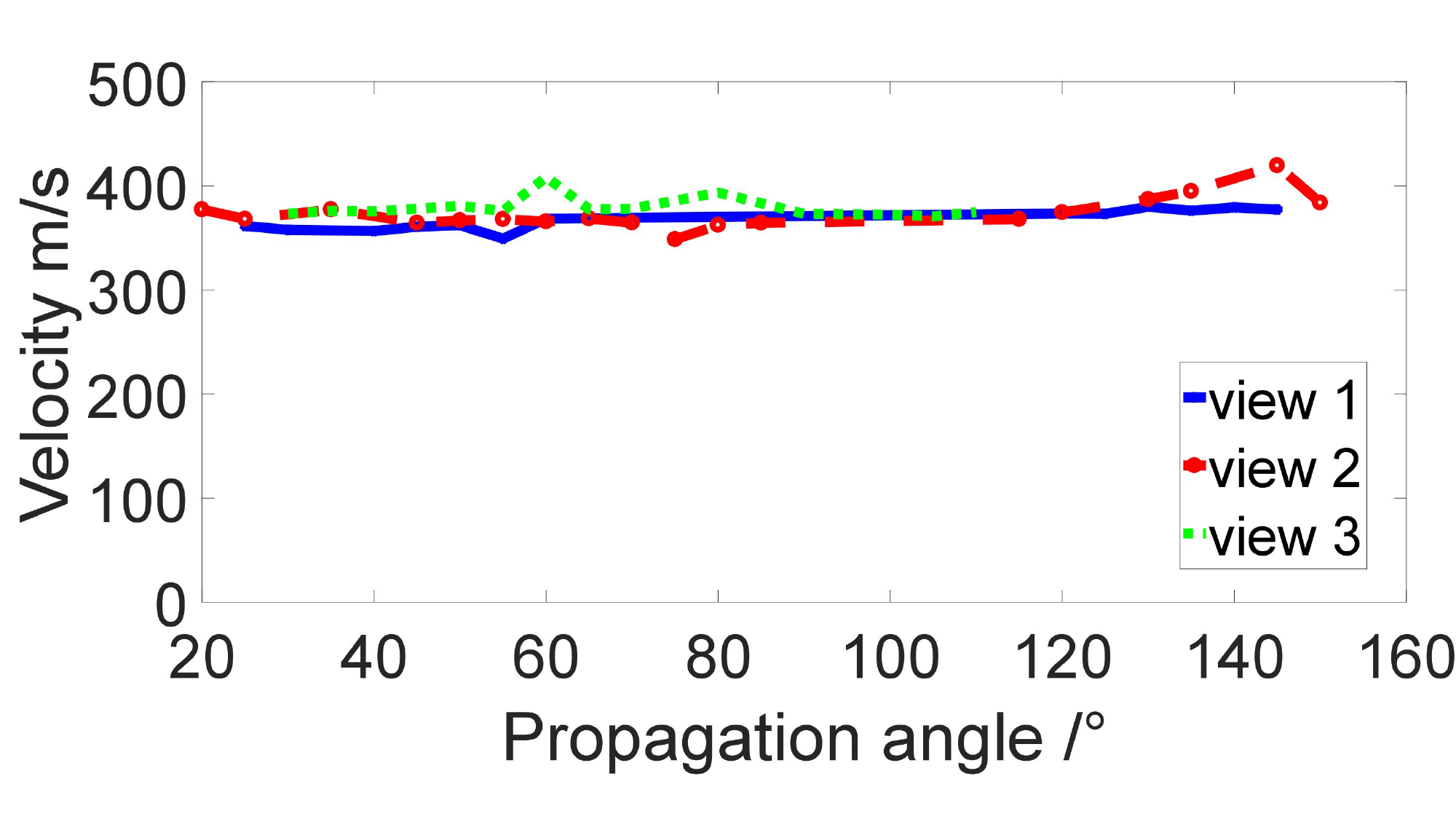}
  \end{subfigure}
  \caption{Radius (top) and velocity (down) measurement results of different propagation angles from multiple views.}
\end{figure}
\subsection{Event-based shock wave measurement and 3D motion field reconstruction}

Based on the calibration and extraction results, the motion parameters of the shock waves in different directions are first measured. Subsequently, the 3D motion field of the shock waves is reconstructed for visualization. Finally, the equivalent charge is reversed from the motion parameters.

\begin{table*}
  \centering
    \caption{Our overall shock wave velocity measurement results and comparison results. * denote our measurement results near the location of the pressure sensor}
  \begin{tabular}{c c c c c c c}
  \hline
  \textbf{Distance} & \multicolumn{2}{c}{$\mathrm{4m}$} & \multicolumn{2}{c}{$\mathrm{6m}$} & \multicolumn{2}{c}{$\mathrm{8m}$} \\
  \hline
  Pressure Sensor ($\mathrm{m/s}$) & \multicolumn{2}{c}{$407.44$} & \multicolumn{2}{c}{$379.42$} & \multicolumn{2}{c}{$362.44$} \\
  Empirical Formula($\mathrm{m/s}$) & \multicolumn{2}{c}{$398.84$} & \multicolumn{2}{c}{$370.76$} & \multicolumn{2}{c}{$360.56$} \\
  \textbf{Ours*} ($\mathrm{m/s}$) & \multicolumn{2}{c}{$\textbf{419.57}$} & \multicolumn{2}{c}{$\textbf{375.02}$} & \multicolumn{2}{c}{$\textbf{362.64}$} \\  
  Absolute Error ($\mathrm{m/s}$) & $+12.13$(P) & $+20.73$(E) & $-4.40$(P) & $+4.26$(E) & $\textbf{+0.20}$(P) & $+2.08$(E) \\    
  Relative Error ($\%$) & $2.98$(P) & $5.20$(E) & $1.16$(P) & $1.15$(E) & $\textbf{0.06}$(P) & $0.58$(E) \\ 
  Ours Average Value ($\mathrm{m/s}$) & \multicolumn{2}{c}{$402.03$} & \multicolumn{2}{c}{$373.38$} & \multicolumn{2}{c}{$360.87$} \\  
  Ours Median ($\mathrm{m/s}$) & \multicolumn{2}{c}{$400.06$} & \multicolumn{2}{c}{$373.22$} & \multicolumn{2}{c}{$362.33$} \\  
  Ours Standard Deviation ($\mathrm{m/s}$) & \multicolumn{2}{c}{$12.38$} & \multicolumn{2}{c}{$13.22$} & \multicolumn{2}{c}{$9.10$} \\  
  \hline
  \end{tabular}
\end{table*}

\subsubsection{Radius and velocity measurement}

In the experiments, the shock wave radius and velocity at different propagation angles under various views are measured, as shown in Fig. 10. To verify the reliability of the results, the uncertainty of the radius measurement is derived, while the pressure sensor measurement results are utilized for velocity comparison. Three free-field pressure sensors are arranged at different angles, positioned $\mathrm{1.5m}$ above the ground and $\mathrm{4m}$, $\mathrm{6m}$, and $\mathrm{8m}$ away from the charge. This arrangement prevents the pressure sensors from obstructing the propagation of shock waves and affecting the results. The overpressure value of the shock wave is converted into velocity for comparison according to Eq.(19). To further demonstrate the reliability of the method, the empirical formula Eq.(20) for shock wave overpressure based on scaled distance $\delta$ is used for comparison. Partial results ($\alpha=55^\circ$) are presented in Fig. 11.

\begin{figure*}
  \centering
  \includegraphics[width=0.95\textwidth,height=!]{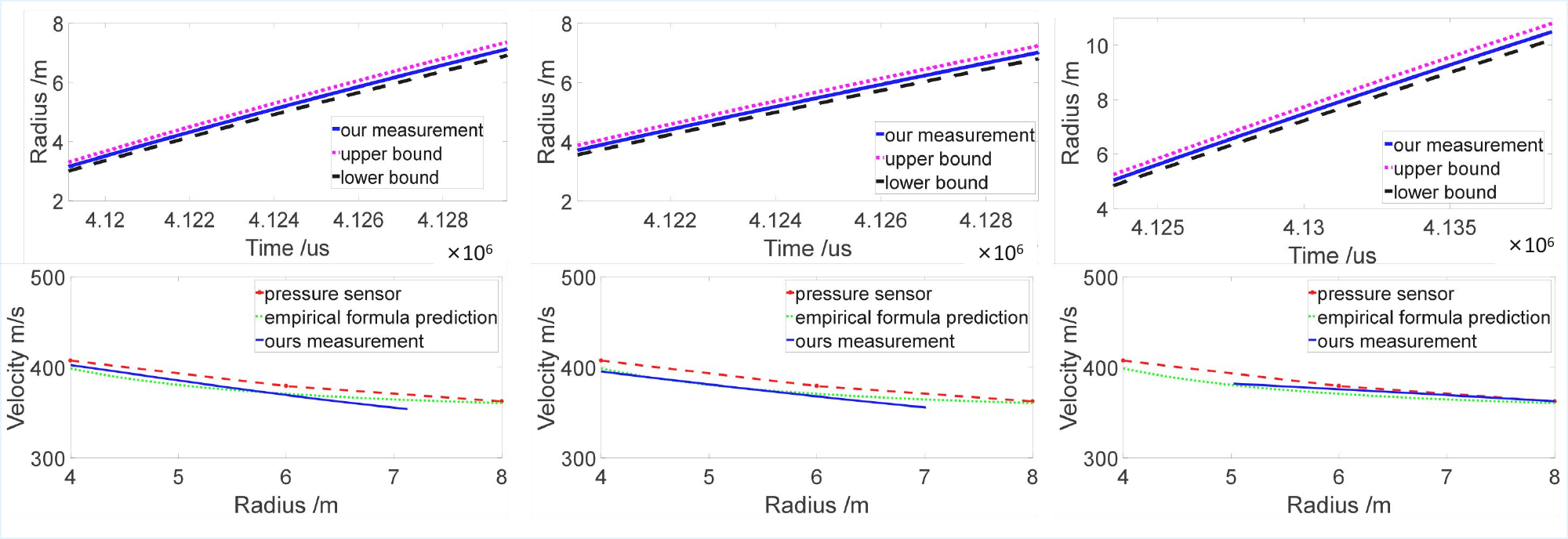}
  \caption{Motion measurement results from multiple views (view 1(left), 2(middle), 3(right)) with $\alpha=55^\circ$. Note: The pressure sensor results are provided for reference only, as their positions do not correspond to the $\alpha=55^\circ$ orientations of the three viewing angles.}
\end{figure*}
\begin{figure*}
  \centering
  \includegraphics[width=\textwidth,height=!]{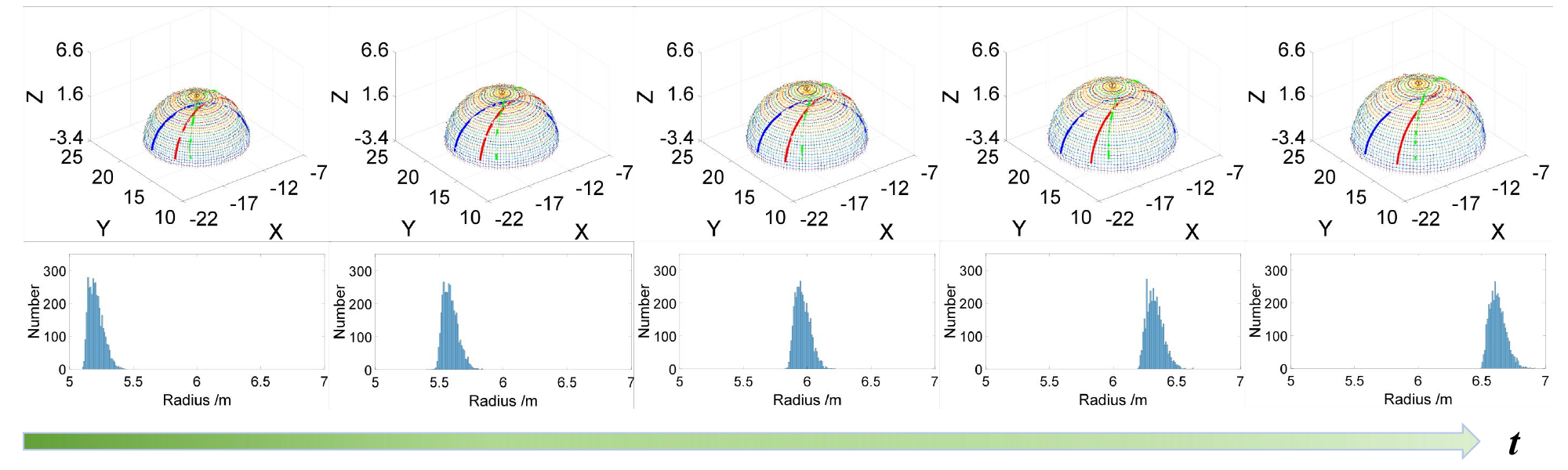}
  \caption{Top: 3D visualization of the shock wave evolution process (modeled with irregular surface). Blue dots (view 1), red dots (view 2), green dots (view 3). Down: the histogram of the radius distribution after uniform sampling from the surface above.}
\end{figure*}

Furthermore, the measurement results of the shock wave motion parameters from three different views at various propagation angles are statistically analyzed, as shown in TABLE II. The measurement results from $4m$, $6m$, and $8m$ away from the charge are utilized for analysis and comparison. We also compared the velocity measurements near the location of the pressure sensor. The experimental results indicate that the shock wave velocity measurement results closely align with those from the pressure sensors and empirical formulas. The minimum error is found to be $0.06\%$, while the maximum error reaches $5.20\%$.

Our polar coordinate encoding is designed to efficiently extract shock waves. Typically, the shock wave region is located more than $100 pixels$ from the explosion center, as the shock waves within the detonation products are not significant. A one-pixel error in estimating the explosion center $(x_b, y_b)$ results in only a minimal shift in shock wave events on the d-t graph when converting to polar coordinates. Therefore, the impact of a one-pixel error on the polar coordinate encoding and shock wave extraction is negligible.

According to Eq.(14), the uncertainty of the radius measurement $\varepsilon _r$ is influenced by the error $\varepsilon _\zeta$, and $\varepsilon _\theta$. The error $\varepsilon _\zeta$ is determined by the measurement accuracy of the RTK, with a value between $\pm \mathrm{0.001m}$. 
 The error $\varepsilon _\theta$ is influenced by the reprojection error $\varepsilon _\pi$ (applied simultaneously to the estimation of blast center coordinates $(x_b, y_b)$ and the imaging of the shockwave front), the focal length error $\varepsilon _f$, the principle point error $\varepsilon _{x_o}$, and $\varepsilon _{y_o}$ from calibration, as well as the extraction error $\varepsilon _a$ of the shock wave. Therefore, the $\varepsilon _r$ is influenced by $\varepsilon _\zeta$, $\varepsilon _f$, $\varepsilon _{x_o}$, $\varepsilon _{y_o}$, $\varepsilon _{a}$ and $\varepsilon _\pi$ through:
\begin{align}
    \left\{
        \begin{aligned}
\varepsilon _r&=G_2(\zeta+\varepsilon _\zeta,\theta+\varepsilon _\theta)-G_2(\zeta,\theta)\\
            \varepsilon _\theta&=G_1(\mathcal{X}')-G_1(\mathcal{X})
        \end{aligned}
    \right.
\end{align} 
where $G_1$ is the function mentioned in Eq.(13), $G_2$ is the function mentioned in Eq.(14). The $\mathcal{X}'$ represents the set of variables after introducing errors with: $x_o'=x_o+\varepsilon_{x_o}$, $y_o'=y_o+\varepsilon_{y_o}$, $x_a'=x_a+\varepsilon_{a_x}+\varepsilon_{\pi_x}$, $y_a'=y_a+\varepsilon_{a_y}+\varepsilon_{\pi_y}$, $x_b'=x_b+\varepsilon_{\pi_x}$, $y_b'=y_b+\varepsilon_{\pi_y}$, $f'=f+\varepsilon_f$. The subscript $x$ of the $\varepsilon_{\pi_x}$, $\varepsilon_{a_x}$ represents pixel errors on horizontal axis, and the same applies to $y$. The angle $\theta$ is always less than $90^{\circ}$, as observations are conducted prior to the shock wave reaching the camera. Furthermore, the $sin$ function within the range of $(0^{\circ}, 90^{\circ})$ is monotonically increasing, indicating that $r$ is positively correlated with $\zeta$ and $\theta$. Based on the calibration results from the left event camera, we select $\mid \varepsilon _f \mid \le \mathrm{20pixels}$, $\mid \varepsilon _{x_o} \mid \le \mathrm{5pixels}$, $\mid \varepsilon _{y_o} \mid \le \mathrm{5pixels}$, $\mid \varepsilon _{a} \mid \le \mathrm{2pixels}$, and $\mid \varepsilon _\pi \mid \le \mathrm{1pixel}$ (Maximum reprojection error $\mathrm{0.88pixels}$) for analysis. Using Eq.(12), the value range of $\overline{CA}$, $\overline{CB}$, and $\overline{AB}$ are determined. Since the $arccos$ function is monotonically decreasing, the upper and lower bound of the error $\varepsilon _r$ are estimated by examining independent variables within the value range, as illustrated in Fig. 11.

\subsubsection{3D motion field reconstruction}

Based on the content of the previous section, the shock wave radius is measured in different propagation directions from multiple views. A common timestamp from these views is then input into the fitted polynomials $\Psi$ (discussed in section III.D.1) determine the radius for each propagation angle. By solving the equations presented in section III.D.2, measurement points from other views are accurately projected into any chosen view, as illustrated in Fig. 13. To facilitate the presentation of results, a time window of $200\upmu\mathrm{s}$ surrounding the selected timestamp is utilized to accumulate events for display. 

\begin{figure}[h]
  \centering
  \includegraphics[width=0.45\textwidth,height=!]{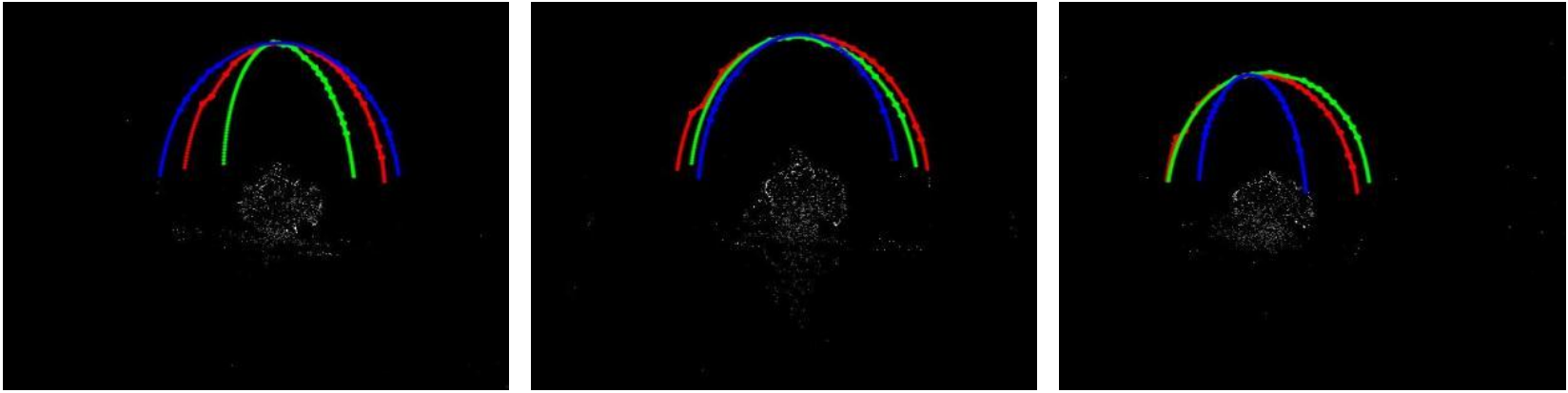}
  \caption{Measurement points from other views are accurately projected into view 1 (left), 2 (middle), and 3 (right). Blue (view 1), red (view 2), green (view 3).}
\end{figure}

Based on the hypothesis that shock waves propagate asymmetrically in various directions, but generally within a certain range, the evolution sequence of 3D coordinates for the measurement points is obtained, as shown in Fig. 12. By utilizing high-temporal resolution optical measurement points, the irregular surface with a known explosion center is used for fitting, enabling detailed visualization of the shock wave propagation. The 3D reconstruction sequence diagram reveals that shock waves propagate outward over time, exhibiting distinct propagation patterns in different directions.

\subsubsection{Charge equivalent reversion}

The charge equivalent is reversed, as shown in TABLE III. Compared to the true value of $\mathrm{600g}$, the minimum error reaches $3.41\%$, while the maximum error can be as high as $17.85\%$. This error is significantly larger than that observed in the measurement of motion parameters, primarily due to error propagation throughout the calculation process. Specifically, according to Eq.(19), the overpressure $P$ is derived by taking the square root of the velocity Mach number $M$. Additionally, as per Eq.(20) and the relationship $\delta=r/\sqrt[3]{W}$, the conversion from explosion equivalence to overpressure $P$ involves the use of a cubic root.
\begin{table}[h]
  \centering
    \caption{Equivalent reversion results of the TNT charge.}
  \begin{tabular}{c c c c}
  \hline  
  Distance from the Blast Center & \textbf{$4m$} & \textbf{$6m$} & \textbf{$8m$} \\
  \hline
  Equivalent Reversion Results $\mathrm{(g)}$ & 657.46 & 707.12 & \textbf{620.48} \\
  Absolute Error $\mathrm{(g)}$ & +57.46 & +107.12 & \textbf{+20.48} \\
  Relative Error $(\%)$ & 9.58 & 17.85 & \textbf{3.41} \\
  \hline
  \end{tabular}
\end{table}

\section{Conclusion}
In this paper, multiple event cameras are employed to measure the motion parameters of conventional explosion shock waves in different directions. The proposed polar coordinate encoding and adaptive ROI region extraction algorithm is utilized to significantly enhance the extraction and retrieval efficiency of shock wave data. Furthermore, a slope-iterative based algorithm is implemented to achieve microsecond-level extraction of shock waves. Based on this, a geometric model of event and shock wave motion parameters and a 3D reconstruction model are developed. These models enable the asymmetric estimation of shock waves to some extent, visualization of the shock wave evolution, and inversion of explosive equivalence. Experimental validation demonstrates that the velocity measurements obtained from multi-view and multi-angle approaches deviate by less than $2.98\%$ from those recorded by pressure sensors. Overall, the proposed approach offers a powerful and innovative solution for precise shock wave characterization and analysis.

\bibliographystyle{IEEEtran}
\bibliography{reference}

\newpage

\begin{IEEEbiography}
	[{\includegraphics[width=1in,height=1.25in,clip,keepaspectratio]{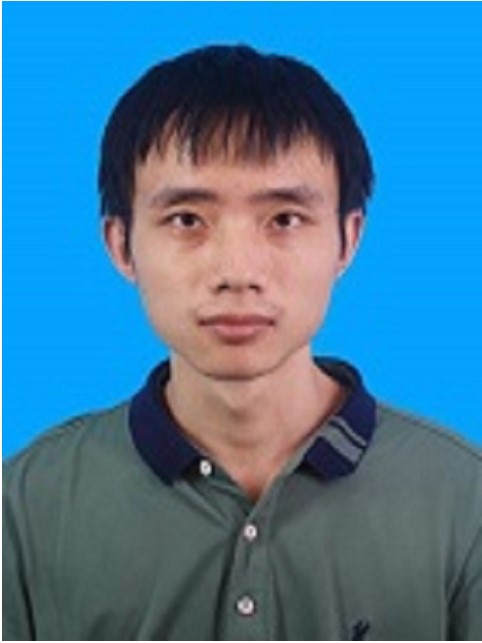}}]{Taihang Lei} 
    received the B.S. degree in computer science and technology from Shanghai University, Shanghai, China, in 2017 and the M.E. degree in computer technology from the Central South University, Changsha, China, in 2020. He is currently pursuing the Ph.D. degree with the College of Aerospace Science and Engineering, National University of Defense Technology, Changsha, China.
    His research interests include computer vision and image processing.
\end{IEEEbiography}
\vspace{-1.5cm}
\begin{IEEEbiography}
	[{\includegraphics[width=1in,height=1.25in,clip,keepaspectratio]{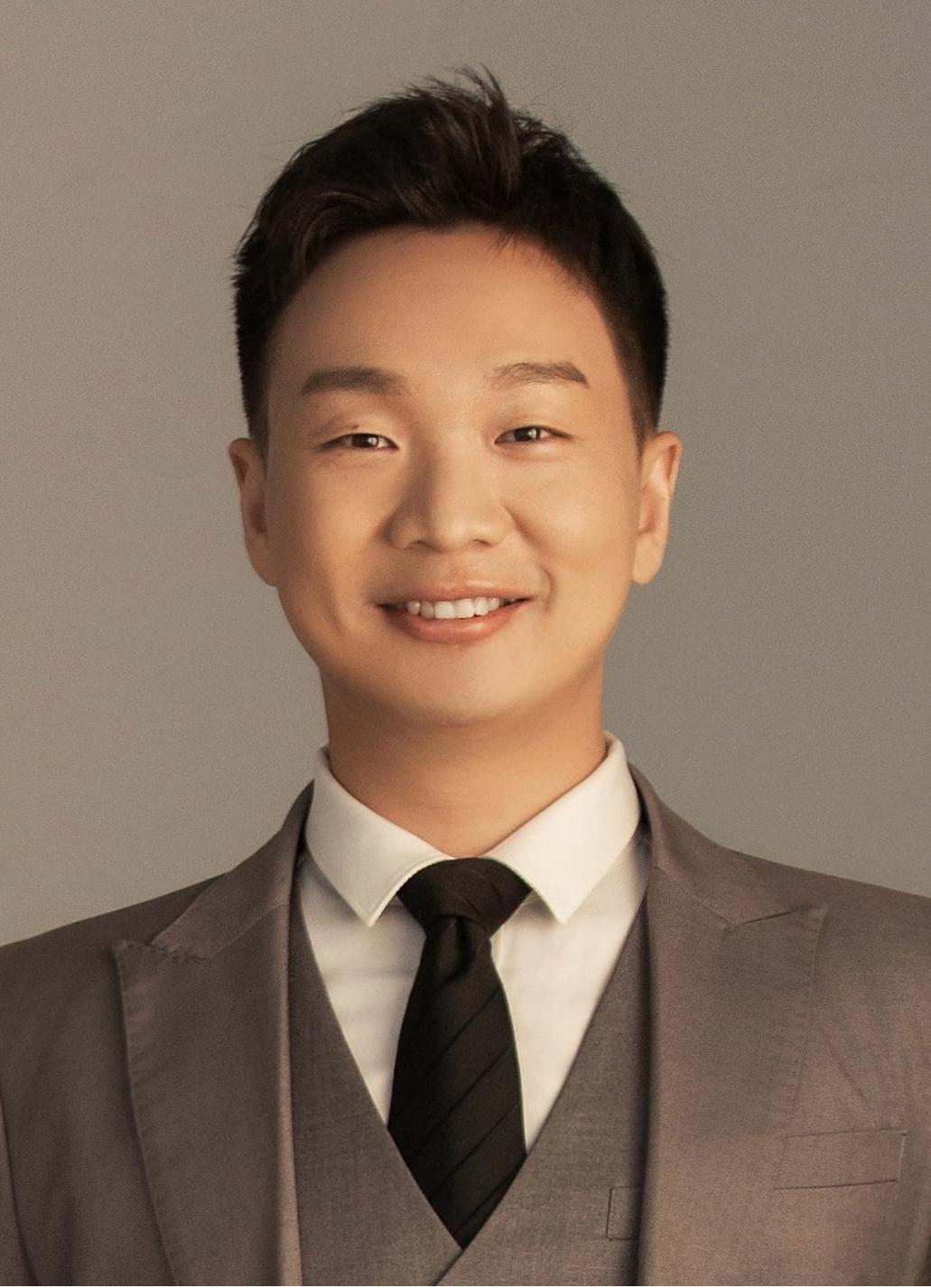}}]{Banglei Guan} 
	received the B.S. degree in geomatics engineering from Wuhan University, Wuhan, China, in 2012 and the Ph.D. degree in aeronautical and astronautical science and technology from the National University of Defense Technology, Changsha, China, in 2018. From 2016 to 2017, he was an Exchange Student with the Graz University of Technology, Graz, Austria.
	He is currently an Associate Professor with the National University of Defense Technology. His research interests include computer vision and photogrammetry.
\end{IEEEbiography}
\vspace{-1.5cm}
\begin{IEEEbiography}
	[{\includegraphics[width=1in,height=1.25in,clip,keepaspectratio]{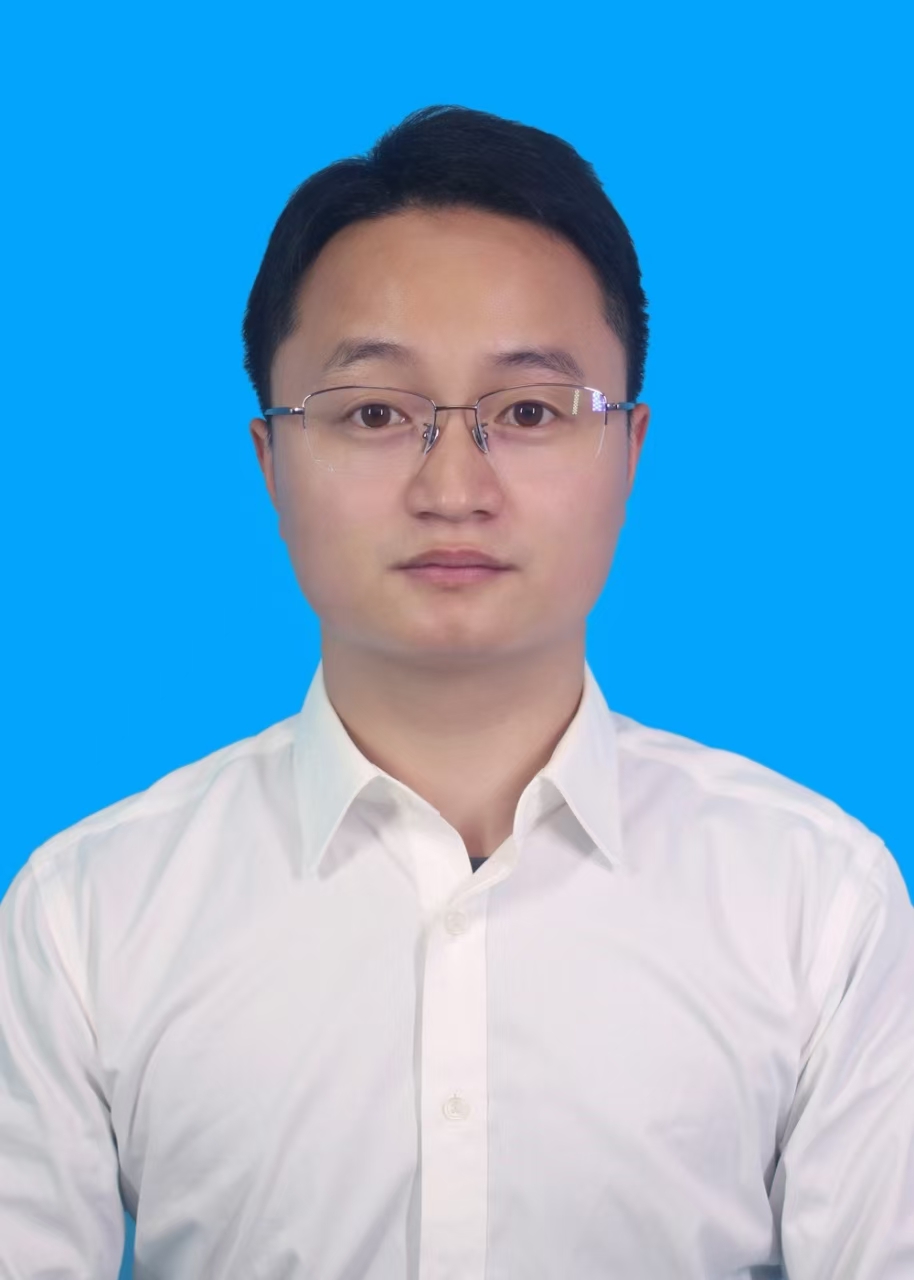}}]{Minzu Liang} 
    elected as an outstanding youth in Hunan Province and a recipient of outstanding doctoral and master's theses in the military. He mainly engages in research on explosion and shock dynamics and serves as a member of the Explosive Group of the National Committee of Explosive Mechanics.
\end{IEEEbiography}
\vspace{-1.5cm}
\begin{IEEEbiography}
	[{\includegraphics[width=1in,height=1.25in,clip,keepaspectratio]{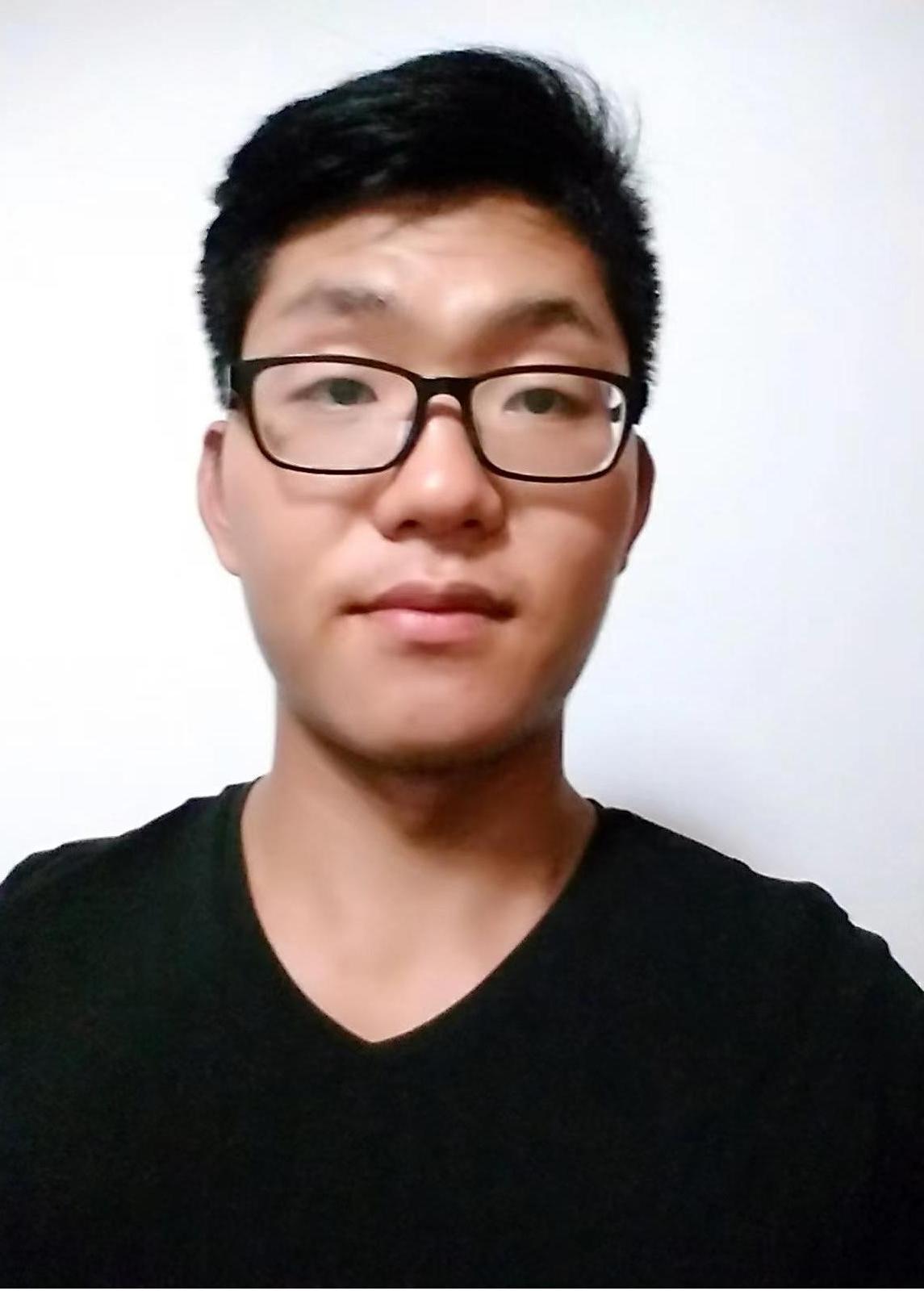}}]{Pengju Sun} 
	received the B.S. degree in geomatics engineering from Wuhan University, Wuhan, China, in 2014 and the M.E. degree from the Institute of Geochemistry, University of Chinese Academy of Sciences, Guiyang, China, in 2022. He is currently pursuing the Ph.D. degree with the College of Aerospace Science and Engineering, National University of Defense Technology, Changsha, China.
	His research interests include computer vision and photogrammetry.
\end{IEEEbiography}
\vspace{-1.5cm}
\begin{IEEEbiography}
	[{\includegraphics[width=1in,height=1.25in,clip,keepaspectratio]{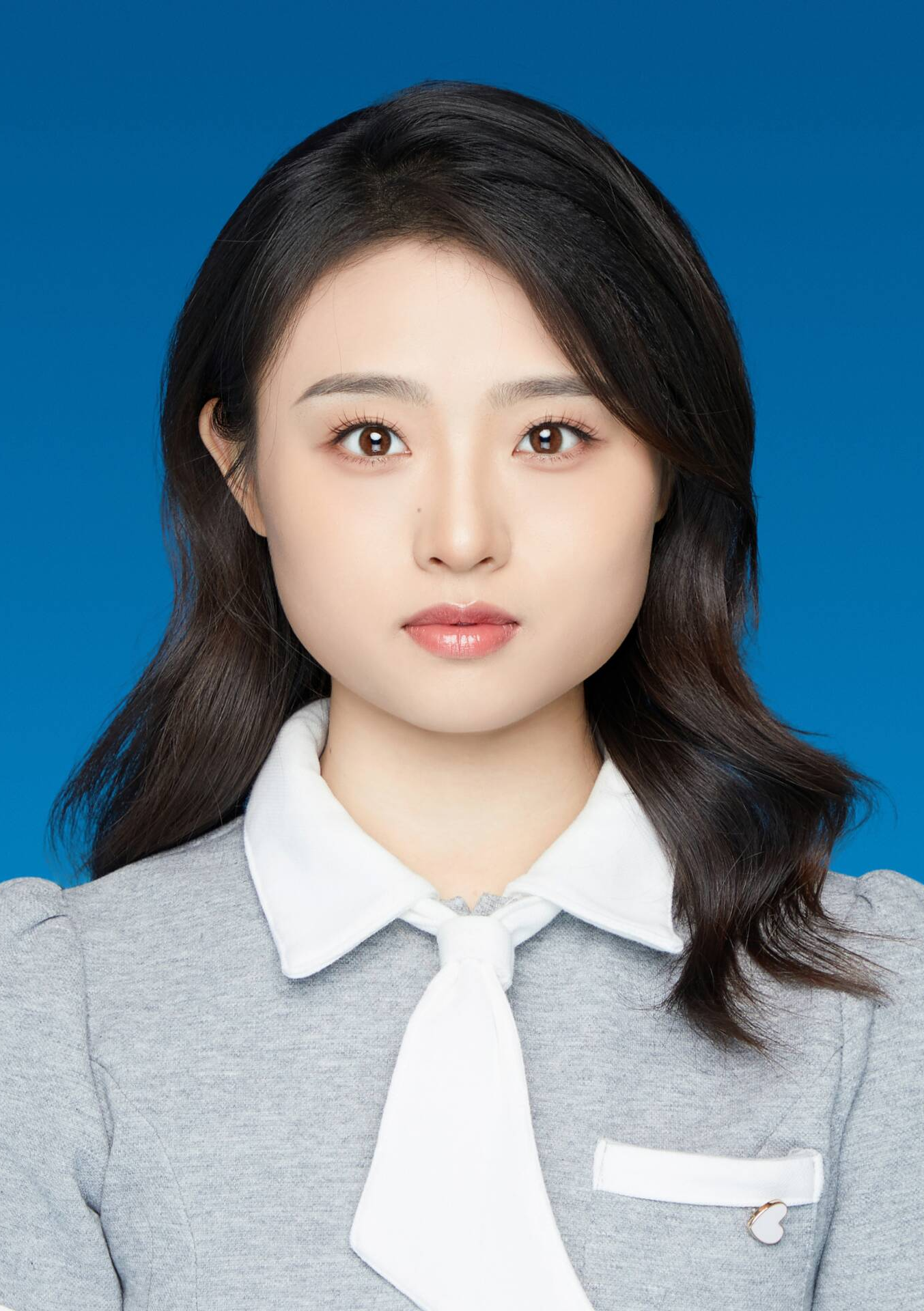}}]{Jing Tao} 
    received the B.S. degree in electronic and information engineering from Hainan University, Haikou, China, in 2017. She is currently pursuing the Ph.D. degree with the College of Aerospace Science and Engineering, National University of Defense Technology, Changsha, China.
    Her research interests include computer vision and image processing.
\end{IEEEbiography}
\vspace{-1.5cm}
\begin{IEEEbiography}
	[{\includegraphics[width=1in,height=1.25in,clip,keepaspectratio]{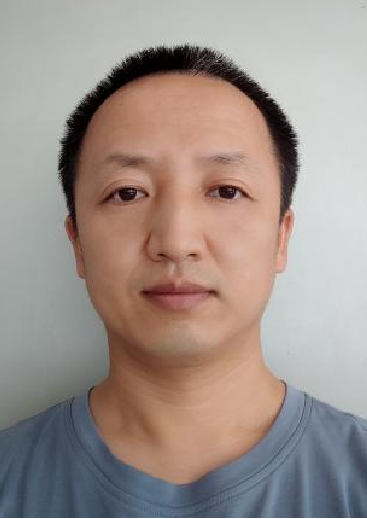}}]{Yang Shang} 
    received the B.E., M.S., and Ph.D. degrees in aerospace science and technology from the National University of Defense Technology, Changsha, China, in 1998, 2000, and 2005 respectively. He is currently a Professor with the College of Aerospace Science and Engineering, Nationa University of Defense Technology. His main research interests include photogrammetry, machine vision, and navigation systems.
\end{IEEEbiography}
\vspace{-1.5cm}
\begin{IEEEbiography}
	[{\includegraphics[width=1in,height=1.25in,clip,keepaspectratio]{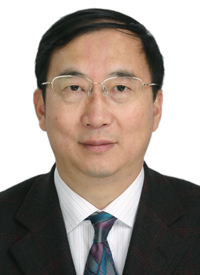}}]{Qifeng Yu} 
	received the B.S. degree in mechanics from Northwestern Polytechnic University, Xian, China, in 1981, the M.S.degree in mechanics from the National University of Defense Technology, Changsha, China, in 1984, and the Ph.D. degree in applied optics from Bremen University, Bremen, Germany, in 1996. He is currently an Academician of CAS and a Professor with the National University of Defense Technology. He has authored three books and published over 100 articles. His main research fields are image measurement, vision navigation, and closerange photogrammetry.
\end{IEEEbiography}

\end{document}